%% file: arxiv.tex
\documentclass[12pt, onecolumn]{article}
\usepackage{arxiv}

\ifpdf \usepackage[pdftex]{graphicx} \pdfcompresslevel=9
\else \usepackage[dvips]{graphicx} \fi 

\usepackage{t1enc}
\usepackage{cite}

\usepackage{hyperref}
\usepackage{microtype}
\usepackage{booktabs}
\usepackage{amsfonts}
\usepackage{graphicx}
\usepackage{amsmath}

\usepackage[dvipsnames]{xcolor}
\usepackage{verbatim}

\newcommand\blfootnote[1]{%
  \begingroup
  \renewcommand\thefootnote{}\footnote{#1}%
  \addtocounter{footnote}{-1}%
  \endgroup
}

\title{FKAConv: Feature-Kernel Alignment for Point Cloud Convolution}
\author{
Alexandre Boulch \\
valeo.ai, Paris
\And
Gilles Puy \\
valeo.ai, Paris
\And
Renaud Marlet\\
valeo.ai, Paris\\
LIGM, Ecole des Ponts, \\Univ Gustave Eiffel,\\ CNRS, Marne-la-Vallée, France
}

\begin{document}

\maketitle
\blfootnote{\color{red}FKAConv code is available on the valeo.ai github: \url{https://github.com/valeoai/FKAConv}}

\input{00_abstract.tex}

\input{01_introduction.tex}
\input{02_related_work.tex}
\input{03_approach.tex}
\input{04_experiments.tex}

\input{05_conclusion.tex}
\clearpage

\bibliographystyle{abbrv}
\bibliography{egbib}

\clearpage
\input{06_supplementary}

\end{document}

%% file: 00_abstract.tex
\begin{abstract}
Recent state-of-the-art methods for point cloud processing are based on the notion of point convolution, for which several approaches have been proposed. In this paper, inspired by discrete convolution in image processing, we provide a formulation to relate and analyze a number of point convolution methods. We also propose our own convolution variant, that separates the estimation of geometry-less kernel weights and their alignment to the spatial support of features. Additionally, we define a point sampling strategy for convolution that is both effective and fast. Finally, using our convolution and sampling strategy, we show competitive results on classification and semantic segmentation benchmarks while being time and memory efficient.
\end{abstract}

%% file: 01_introduction.tex
\section{Introduction}

Convolutional Neural Networks (CNNs) have been a breakthrough in machine learning for image processing~\cite{ciresan2011cnn,krizhevsky2012alexnet}.
The discrete formulation of convolution allows a very efficient processing of grid-structured data such as images in 2D or videos in 3D.
Yet a number of tasks require processing unstructured data such as point clouds, meshes or graphs, with application domains such as autonomous driving, robotics or urban modeling.
However discrete convolution does not directly apply to point clouds as 3D points are not usually sampled on a grid.

The most straightforward workaround is to voxelize the 3D space to use discrete CNNs~\cite{maturana2015voxnet}. However, as 3D points are usually sampled on a surface, most of the voxels are empty.
For efficient large-scale processing, a sparse formulation is thus required~\cite{riegler2017octnet,zhou2018voxelnet}.
Other deep learning approaches generalize convolution to less structured data, such as graphs or meshes~\cite{scarselli2008graph,bronstein2017geometric}, but applying them to point clouds requires addressing the issue of sensible graph construction first.

Deep-learning techniques that directly process raw data have been developed to overcome the problem of point cloud pre-processing~\cite{qi2017pointnet,wang2018deep}.
Just as for structured data, such networks are usually designed as a stack of layers and are optimized using stochastic gradient descent and back-propagation. Key issues when designing these networks include speed and memory efficiency.

In this context, we propose a new convolution method for point cloud processing. 
It is a mixed discrete-continuous formulation that disentangles the geometry of the convolution kernel and the spatial support of the features: using a geometry-less kernel domain, we stick to a discrete convolution scheme, which is efficient and has been successful on grid data; the spatial domain however keeps its continuous flavor, as point clouds are generally sampled on manifolds.

Our contributions are the following: (1)~we provide a formulation to relate and analyze existing point convolution methods; (2)~we propose a new convolution method (FKAConv) that explicitly separates the estimation of geometry-less kernel weights and their alignment to the spatial support of features; (3)~we define a point sampling strategy for convolution that is both efficient and fast; (4)~experiments on large-scale datasets for classification and semantic segmentation show we reach the state of the art, while being memory and time efficient.

%% file: 02_related_work.tex
\section{Related work}

\textit{Projection in 2D.}
Some methods project the point cloud in a space suitable for using standard discrete CNNs.
2D CNNs have been use for 3D data converted as range images~\cite{gupta2014learning,long2015fully} or viewed 
from virtual viewpoints~\cite{su2015multi,boulch2017snapnet,lawin2017deep}. 
As neighboring points in the resulting image can be far away in 3D space, 2D CNNs often fail to capture well 3D relations.
2D CNNs can also be applied locally to point-specific neighborhoods by projecting data on the tangent plane~\cite{tatarchenko2018tangent}; the result is then highly dependant on the tangent plane estimation.
Other approaches use a volumetric data representation, such as voxels~\cite{maturana2015voxnet,roynard2018classification,qi2016volumetric,wu20153d}.
These approaches however suffer from encoding mostly empty volumes, calling for sparsity handling, e.g., with octree-based 3D-CNNs \cite{riegler2017octnet} or sparse convolution~\cite{graham2014spatially,graham20183d}.

\textit{Graph convolution, geometric deep learning.}
Graph Neural Networks (GNNs) \cite{scarselli2008graph,bronstein2017geometric} extend neural networks to irregular structures (not on a grid), using edges between nodes for message passing~\cite{gilmer17,li2015gated} or defining convolution in the spectral domain~\cite{bruna14,defferrard2016convolutional,kipf17}. Point convolution using GNNs requires first explicitly building a graph from the point cloud~\cite{qi20173d}.
To scale to large point clouds, SPG~\cite{landrieu2018large} defines a graph over nodes corresponding to point segments.
In contrast, our approach directly applies to the raw point cloud, with no 
predefined relation between points, somehow making point association as part of the method.

\textit{MLP processing.}
PointNet~\cite{qi2017pointnet} directly processes point coordinates with a multi-layer perceptron (MLP), gathering context information with a permutation-invariant max-pooling.
PointNet++~\cite{qi2017pointnet++} and So-Net~\cite{li2018so} reduce the loss of local information due to subsampling with a cascade of 
MLPs at different scales.

\textit{Point convolution.}
A first line of work considers an explicit spatial location for the kernel, in the same space as the point cloud.
Kernel elements can be located on a regular grid (voxels)~\cite{hua2018pointwise}, at the vertices of a polyhedron~\cite{thomas2019kpconv} or randomly sampled and optimized at training~\cite{boulch2020convpoint}.
In KPConv~\cite{thomas2019kpconv}, an adjustment of the kernel locations may also be predicted at test time to better fit the data.

Another type of approaches models kernel locations implicitly.
The kernel can be a family of polynomials like in SpiderCNN~\cite{xu2018spidercnn}, or it can be estimated with an MLP, like in PCCN~\cite{wang2018deep}, RSConv~\cite{liu2019relation} or PointConv~\cite{wu2019pointconv}. The weights of the input features are then directly estimated based on the local geometry of points.
In contrast, we learn the weights of a discrete kernel and, at inference time, we only estimate the spatial relation between the kernel and input points.
PointConv~\cite{wu2019pointconv} reweights the input features based on local point densities. Our method reaches state-of-the-art performances without the need of such a mechanism.

Finally, PointCNN~\cite{li2018pointcnn} shares apparent similarities with our work as one of its main components is the estimation of a matrix, that actually differs from ours.
Besides, geometric information in~\cite{li2018pointcnn} is lifted to the feature space and used as additional features.
Our work shows it is sufficient to use the geometry only for features-kernel alignment, mimicking the discrete convolution on a regular grid.

Our approach lies in between these lines of work. 
On the one hand, our kernel weights are explicitly modeled as in~\cite{hua2018pointwise,boulch2020convpoint,thomas2019kpconv}, which gives a discrete flavor to our method; on the other hand, we estimate a transformation of input points to apply the convolution as in~\cite{wang2018deep,liu2019relation,wu2019pointconv}, which operates in the continuous domain, avoiding kernel spatialization.
The key is that, contrary to fully-continuous approaches that re-estimate at inference time how to weigh given sets of points to operate the convolution, we estimate separately a kernel while learning, and we predict the relation between the kernel and input points while testing. Besides, we perform the convolution with a direct matrix multiplication rather than getting indirectly results from a network output. This separation and the explicit matrix multiplication (outside the network) allows a better learning of kernel weights and spatial relations, without the burden and inaccuracy of estimating their composition, resulting in a time and memory efficient method.

\relax{
\textit{Point sampling.}
Like PointNet~\cite{qi2017pointnet}, several methods maintain point clouds at full resolution during the whole processing~\cite{yang2019modeling,liu2019dynamic,liu2019point}.
These methods suffer from a high memory cost, which requires to either limit the input size~\cite{qi2017pointnet,yang2019modeling}, split the input into parts~\cite{liu2019dynamic}, or use a coarse voxel grid \cite{liu2019point}.
Other approaches \cite{qi2017pointnet++,li2018pointcnn,boulch2020convpoint}, as ours, use an internal sub-sampling of the point cloud.
The choice of
sampled points forming a good support 
is a key step for this reduction.
\textit{Furthest point sampling} (FPS) \cite{qi2017pointnet++}, where points are chosen iteratively by selecting the furthest point from all the previously picked points at each iteration, yields very good performance but is slow and its performance depends on the initialization.
In \cite{yang2019modeling}, point sampling is based on a learned attention, which induces a high memory cost.
Our sampling strategy, based on the quantization of the 3D space, ensures a good sampling of the space, like FPS, and is fast and memory efficient.
}

%% file: 03_approach.tex
\section{A general formulation of point cloud convolution}

We base our convolution formulation on the discrete convolution used in image or voxel grid processing.
The formulation is general enough to cover a wide range of state-of-the-art convolution methods for point clouds, and to relate them.

\subsubsection{Discrete convolution.}

\begin{figure}[t]
    \centering
    \begin{tabular}{c@{\hspace{0.2cm}}c@{\hspace{0.2cm}}c}
    \includegraphics[width=0.26\linewidth]{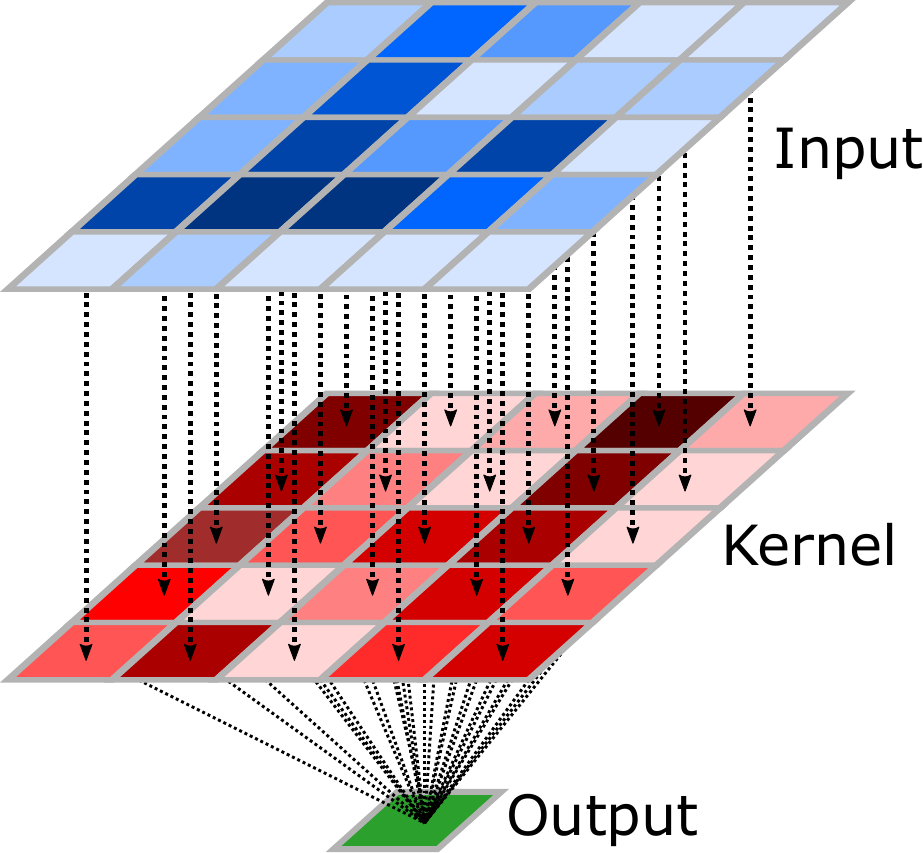}&
    \includegraphics[width=0.31\linewidth]{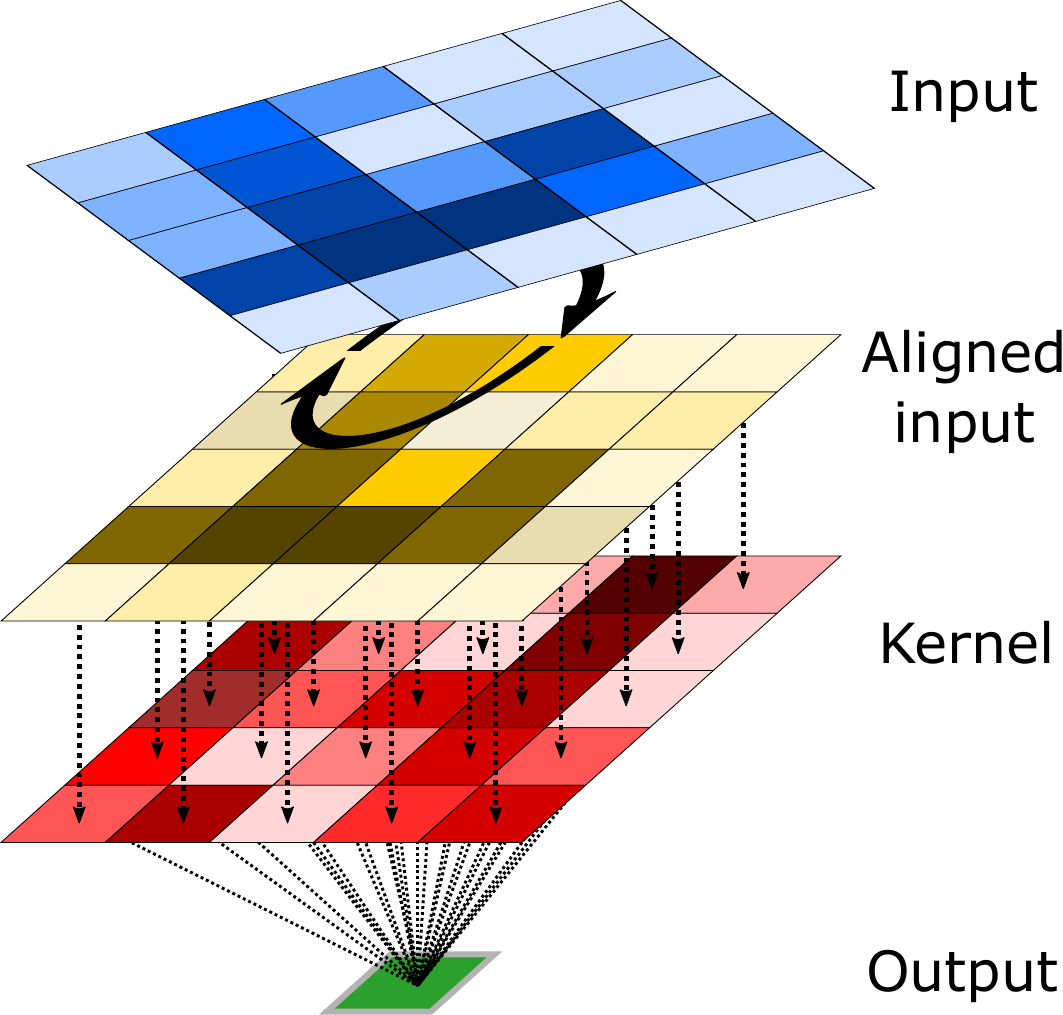}&
    \includegraphics[width=0.38\linewidth]{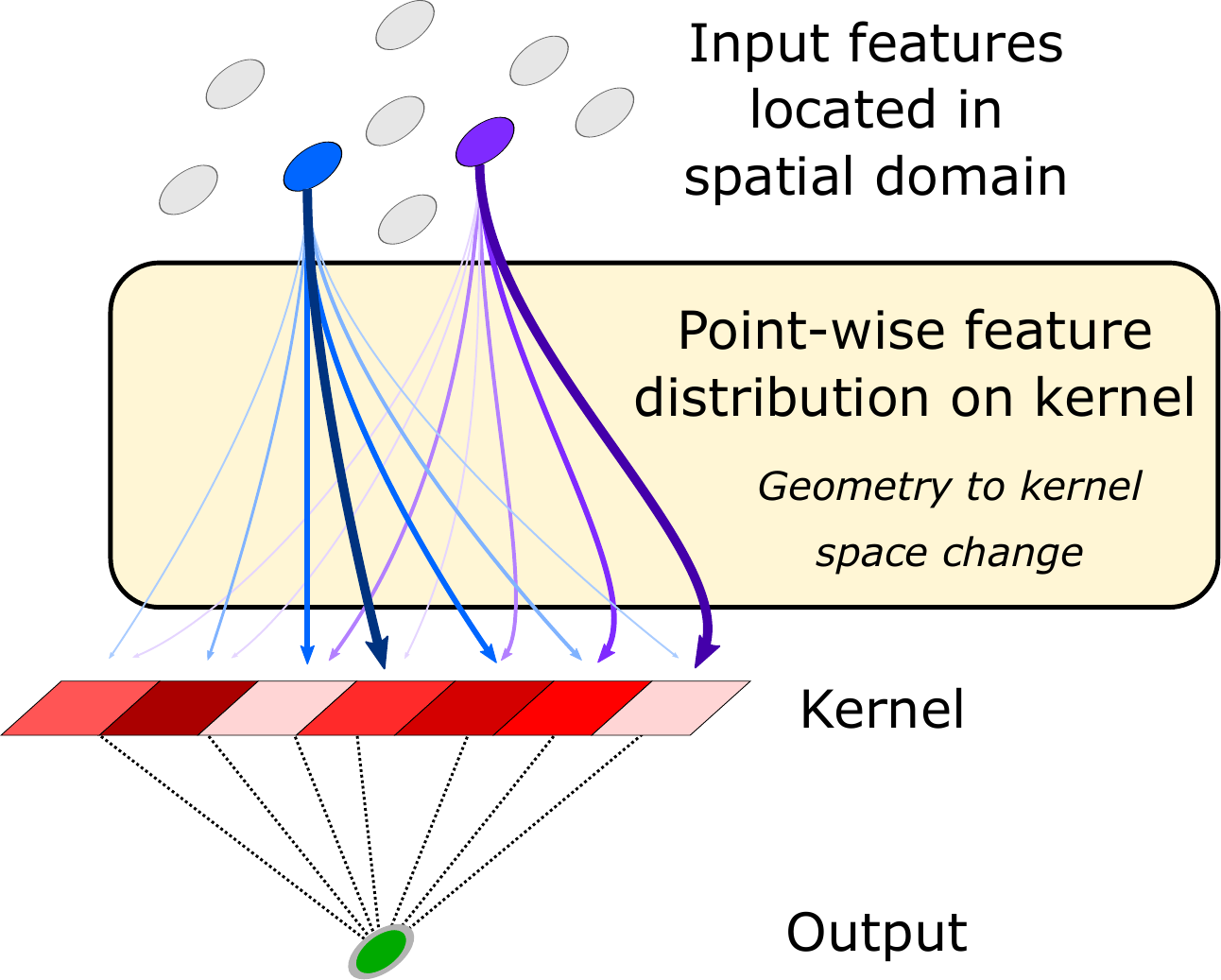}\\
    {\scriptsize (a) Discrete conv.\ aligned}&
    {\scriptsize (b) Discrete conv.\ misaligned}&
    {\scriptsize (c) FKAConv point convolution} \\
    \end{tabular}
    \caption{Kernel-input alignment for grid inputs (a,b) and point clouds (c).}
    \label{fig:alignment}
\end{figure}

Let $F$ be the dimension of the input feature space, $d$ the spatial dimension (e.g., 2 for images, 3 for voxel grids), $\mathbf{K}$ the convolution kernel, and $\mathbf{f}$ the input features. The classical discrete convolution, noted $\mathbf{h}$, is:
\begin{equation}
\label{eq:discrete}
    \mathbf{h}[n] = \sum_{f \in \{1, \dots, F\}} \ \sum_{m \in \{-M/2, \dots, M/2\}^d} \mathbf{K}_f[m] \:\mathbf{f}_f[n+m],
\end{equation}
where $M^d$ is the grid kernel size, $f$ indexes the feature space, $n$ is the spatial index, and $\mathbf{K}_f[m]$ and $\mathbf{f}_f[n+m]$ are scalars.
Defining vectors $\mathbf{K}_f = ( K_f[m] , \, m \in \{-M/2, \dots, M/2\}^d \})$ and $\mathbf{f}_f(n) = ( \mathbf{f}_f[n+m] , \, m \in \{-M/2, \dots, M/2\}^d )$, we can highlight the separation between the kernel space ($\mathbf{K}$) and the feature space ($\mathbf{f}$):
\begin{equation}
    \label{eq:discrete2}
    \mathbf{h}[n] = \sum_{f \in \{1, \dots, F\}} \underbrace{\mathbf{K}_f^\top \vphantom{\mathbf{K}_f^\top}}_{\text{Kernel space}} \: \underbrace{\mathbf{f}_f(n)   \vphantom{\mathbf{K}_f^\top}}_{\text{Feature space}}.
\end{equation}
The kernel $\mathbf{K}_f$ and the features $\mathbf{f}_f(n)$ are perfectly aligned: the grid index $m$ associates a kernel element $\mathbf{K}_f[m]$ with a single input element $\mathbf{f}_f[n+m]$ (Fig.\,\ref{fig:alignment}(a)).

\subsubsection{Point convolution.}

To generalize this discrete convolution to point clouds, we first consider a hypothetical misalignment between the feature and kernel spaces, assuming the feature grid is rotated with respect to the kernel grid (Fig.\,\ref{fig:alignment}(b)), thus obfuscating the correspondence between kernel elements and feature elements.
Yet, provided that the rotation matrix $\mathbf{A} \in \mathbb{R}^{M^d} \times \mathbb{R}^{M^d}$ is known, the correspondences can be recovered by rotating the support points of features:
\begin{equation}
    \label{eq:continuous}
    \mathbf{h}[n] = \sum_{f \in \{1, \dots, F\}} \mathbf{K}_f^\top \mathbf{A} \:\mathbf{f}_f(n).
\end{equation}
This equation actually holds in a more general setting, with an arbitrary linear transformation between the feature space and the kernel space; $\mathbf{A}$~is then the \emph{alignment matrix} that associates the feature values to the kernel elements.

The discrete convolution on a regular grid becomes a particular case of Equation~\eqref{eq:continuous}, with $\mathbf{A} = \mathbf{I}_{M^d}$, the identity matrix.
In the case of a point cloud, $\mathbf{f}_f(n)$ is the feature associated to the point at spatial location $n$, typically computed on 
a neighborhood $\mathbf{N}[n]$. 
These features are generally not grid-aligned.
But Equation~\eqref{eq:continuous} can still apply, provided we can estimate an alignment matrix $\mathbf{A}$ that distributes each input point onto the kernel elements (Fig.\,\ref{fig:alignment}(c)).

In this context, a fixed matrix $\mathbf{A}$ is suboptimal as it cannot cope well with both a regular grid ($\mathbf{A} = \mathbf{I}_{M^d}$) and arbitrary point configurations in a point cloud. 
$\mathbf{A}$ thus has to be a function of the input points, which in practice have to be limited to neighbors $\mathbf{N}[n]$ at location $n$.
The convolution becomes:
\begin{equation}
\label{eq:mixed}
    \mathbf{h}[n] = \sum_{f \in \{1, \dots, F\}} \mathbf{K}_f^\top \mathbf{A}(\mathbf{N}[n]) \: \mathbf{f}_f(n).
\end{equation}
It is a mixed discrete-continuous formulation: $\mathbf{K}_f$ and $\mathbf{f}_f(n)$ have a discrete support and continuous values, while $\mathbf{A}(\mathbf{N}[n])$ provides a continuous mapping.

\subsubsection{Analysis of exiting methods.}

This formulation happens to be generic enough to describe a range of existing methods for point convolution \cite{su2018splatnet,thomas2019kpconv,boulch2020convpoint,wang2018deep,li2018pointcnn}. 

\textit{Using spatial kernel points.}
The most common approach to discrete convolution on a point cloud assigns a spatial point to each kernel element.
The distribution of features on kernel elements is then based on the distance between kernel points and points in $\mathbf{N}[n]$, corresponding to an association matrix $\mathbf{A}$ invariant by rotation.
A simple method would be to assign the features to the nearest kernel point, but it is unstable as a small perturbation in the point position may result in a different kernel point attribution.
A workaround is to distribute the input points to several close kernel points. In SplatNet~\cite{su2018splatnet}, an interpolation distribute points onto the kernel space.
However, this handcrafted assignment is arbitrary and heavily relies on the geometry of kernel points.
KPConv~\cite{thomas2019kpconv} chooses to distribute the input points over all the neighboring kernel points, with a weight inversely proportional to their distance to kernel points.
Moreover, KPConv allows deformable kernels, for which local shifts of kernel points are estimated, offering more adaptation to input points. Yet, this handcrafted distribution is still arbitrary and still relies on the geometry of kernel points.
ConvPoint~\cite{boulch2020convpoint} randomly samples the kernel points, and their position is learned along with an assignment function~$\mathbf A$, with an MLP is applied to the kernel points represented in the coordinate system centered on the input points.
All these methods~\cite{su2018splatnet,thomas2019kpconv,boulch2020convpoint} raise the issue of defining and optimizing the position of kernel points.

\textit{Feature combination and geometry lifting.}
In PointCNN~\cite{li2018pointcnn}, geometric information is extracted with an MLP$_\delta$, parameterized by $\delta$, and concatenated with the input features to create mixed spectral-geometric features.
The summands in Equation~\eqref{eq:mixed} become $\mathbf{K}^\top\mathbf{A}(\mathbf{N}[n])  [\mathbf{f}_f(n), \text{MLP}_\delta(\mathbf{N}[n])]$.

\textit{Joint estimation of \,$\mathbf{K}^\top \mathbf{A}(\mathbf{N}[n])$.}
In fully implicit approaches~\cite{wang2018deep,boulch2020convpoint,thomas2019kpconv}, MLPs are used to directly estimate the weights $\mathbf{W}(n)$ to apply to input features $\mathbf{f}_f(n)$, i.e., not separating $\mathbf{W}(n)$ into a product $\mathbf{K}^\top\times\mathbf{A}(\mathbf{N}[n])$, and thus mixing estimations in the spatial and feature domains.

\textit{Kernel separation.}\label{sec:kernelsep}
As acknowledged by the authors of PCCN themselves \cite{wang2018deep}, a direct estimation of $\mathbf{W}(n) \,{=}\, \mathbf{K}^\top \mathbf{A}(\mathbf{N}[n])$ is too computationally expensive to be used in practice.
Instead, they resort to an implementation which falls into our formulation: for $N_{\text{out}}$ output channels, they consider $N_{\text{out}}$ parallel convolution layers with a size-$1$ kernel, corresponding to using a different $\mathbf{A}$ for each filter.
In PointCNN~\cite{li2018pointcnn}, the extra features generated by the geometry lifting induce a larger kernel (1/4 more weights with default parameters~\cite{li2018pointcnn}), thus an increased memory footprint. To overcome this issue, \cite{li2018pointcnn}~also chooses to factorize the kernel as the product of two smaller matrices. Notice that this kernel separation trick can be implemented in any method that uses explicit kernel weights. Although we could as well, we do not need to resort to that trick in our method.

\begin{figure*}[t]
    \centering
    \includegraphics[width=\linewidth]{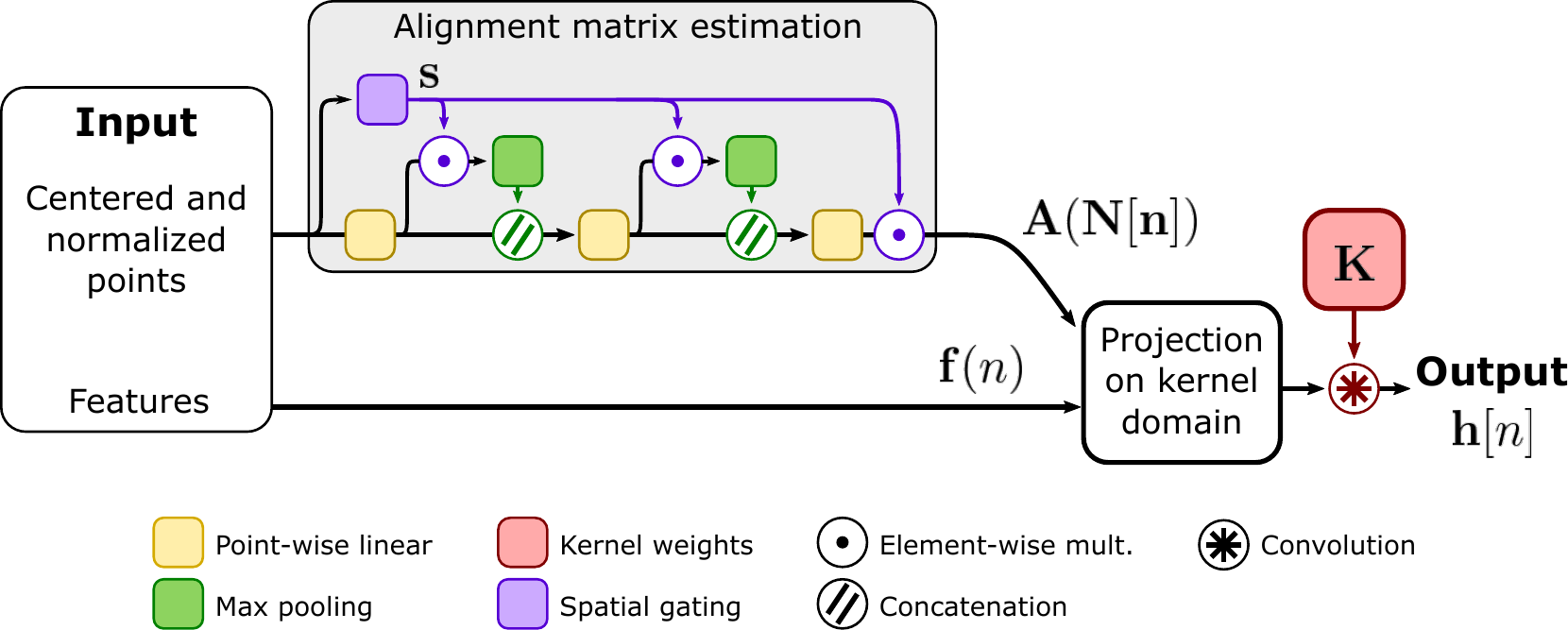}
    \vspace{-4mm}
    \caption{FKAConv convolutional layer.}
    \label{fig:conv_layer}
    \vspace{-2mm}
\end{figure*}

\section{Our method: estimating a feature-kernel alignment}\label{sec:fkaconv}

Our own convolution method is also based on Equation~\eqref{eq:mixed}. However, contrary to preceding approaches, we do not use kernel points. Instead, we estimate a soft alignment matrix $\mathbf A$ based on the coordinates of neighboring points in $\mathbf{N}[n]$. Our convolutional layer is illustrated on Figure~\ref{fig:conv_layer}.

\textit{Neighborhood normalization.}
To be globally invariant to translation, all coordinates of the points of $\mathbf{N}[n]$ are expressed in the local coordinates system of $n$.
This is particularly important for scene segmentation: the network should behave the same way for similar objects at different spatial locations.
Please notice that it is not the case in RSConv~\cite{liu2019relation} where absolute coordinates are used, making it appropriate for shape analysis, not for scene processing.

$\mathbf{N}[n]$ is typically defined as the $k$-nearest neighbors ($k$-NNs) of $n$, or as all points in a ball around $n$. Both definitions have pros and cons. Using $k$-NNs is relatively fast, but the radius of the encompassing ball is (potentially highly) variable.
As observed in~\cite{thomas2019kpconv}, it may degrade spatial consistency compared to using a ball with a fixed radius. But searching within a radius is slower and yields (potentially widely) different numbers of neighbors, requiring strategies to deal with variable sizes, e.g., large tensor sizes and size tracking as in~\cite{thomas2019kpconv}.

We propose an intermediate approach based on the $k$ nearest neighbors, with a form a rescaling. As opposed to~\cite{qi2017pointnet++,li2018pointcnn,boulch2020convpoint}, we do not normalize the neighborhood to a unit sphere regardless of its actual size in the original space.
We estimate a normalization radius $r_t$ of the neighborhood at the layer level using the exponential moving average of Eq.\,\eqref{eq:radius} computed at training time, where $t$ is the update step, $m$ is a momentum parameter and $\hat{r}_t$ is the average neighborhood radius of the current batch.
Let $\mathbf{q}$ be the support point associated to $n$, and $\mathbf{p}_i$ the $i$-th point of $\mathbf{N}[n]$. 
The points $(\hat{\mathbf{p}}_i)_i$ actually used for the estimating $\mathbf A$ are the points $(\mathbf{p}_i)_i$ centered and normalized using $\mathbf{q}$ and $r_t$ as follows:
\begin{eqnarray}
r_{t} &=& \hat{r}_t * m + r_{t-1} * (1-m), \label{eq:radius}\\
\hat{\mathbf{p}}_i &=& (\mathbf{p}_i-\mathbf{q}) / r_{t}. \label{eq:localcoord}
\end{eqnarray}
At inference time, this normalization ensures that all neighborhood are processed at the same scale while on average, neighborhoods are mapped to the unit ball.

\textit{Gating mechanism on distance to support point.}
While solving the problem of the neighborhood scale, this normalization strategy does not prevent points far away from the support point (the neighborhood center) to influence negatively the result.
One could use hard-thresholding on the distance based on the estimated normalization radius $r$ to filter these points, but this approach may cut too much information from the neighborhood, particularly in the case of high variance in neighborhood radii.
Instead, we propose a gating mechanism to reduce, if needed, the effect of such faraway points.
Given $(\hat{\mathbf{p}}_i)_i$ as defined in Equation~\eqref{eq:localcoord}, the spatial gate weight $\mathbf{s} = (s_i)_i$ satisfies
\begin{equation}
    s_i = \sigma(\beta - \alpha ||\hat{\mathbf{p}}_i||_2),
\end{equation}
where $\sigma(\cdot)$ is the sigmoid function, $\beta$ is the cutoff distance (50\% of the maximal value) and $\alpha$ parametrize the slope of the transition between $0$ (points filtered out) and $1$ (points kept).
Both $\alpha$ and $\beta$ are learned layer-wise.

\textit{Estimation of $\mathbf{A}$.}
As underlined in \cite{qi2017pointnet}, a point cloud is invariant by point permutation: changing the order of points should not change the point cloud properties.
Hence, the product $\mathbf{A} \: \mathbf{f}_f(n)$ must be invariant by permutation of the inputs.
This can be achieved by estimating independently each line $\mathbf{A}_j$ of the matrix $\mathbf{A}$ using only the corresponding point $\mathbf{p}_j \in \mathbf{N}[n]$, with an MLP shared accross all points.
But it does not take the neighborhood into account, and thus may ignore useful information such as the local normal or curvature.
To address point permutation invariance, PointNet~\cite{qi2017pointnet} uses a max-pooling operation.
Likewise, we use a three-layer point-wise MLP with max-pooling after the first two layers.
To reduce the influence of outliers, max-pooling inputs are weighted with~$\mathbf{s}$.
The output is then concatenated to the point-wise features and given as input to the next fully-connected layer. This series of computations is illustrated in Figure~\ref{fig:conv_layer} in the block called ``alignment matrix estimation.''

\section{Efficient point sampling with space quantization}

Networks architectures for point cloud processing operates at full resolution through the entire network~\cite{wang2018deep}, or have an encoder/decoder structure~\cite{li2018pointcnn,boulch2020convpoint} similar to networks used in image processing, e.g., U-Net~\cite{ronneberger2015u}.
While the former maintain a maximum of information through the network, the later are usually faster as convolutions are applied to smaller point sets.
However, decreasing the size of the point cloud requires to select \textit{the support points}, i.e., the points at the center of the neighborhoods used in the convolution.

PointNet++~\cite{qi2017pointnet++} introduces farthest point sampling, an iterative sampling procedure where the next sampled point is the farthest from the already picked points.
The main advantage of this sampling is to ensure a somewhat spatially uniform distribution which favors extreme points (e.g., at wing extremities for planes) as they are usually important for shape recognition.
However, it requires to keep track of distances between all pairs of points, which is costly and increases the computation time, in particular when dealing with large point clouds.

In ConvPoint~\cite{boulch2020convpoint}, the point-picking strategy only takes into account seen and unseen  
points, without distance consideration. Points are randomly picked among points that were not previously seen (picked points and points in the neighborhood of these points).
While being much faster than farthest point sampling, it appears to be less efficient (see experiments in Section~\ref{sec:expe}).
In particular, the sampling is dependent of the neighborhood size: the sampling is done outside
the neighborhoods of the previously picked support points. A very small neighborhood size reduces the method to a pure random sampling.

\begin{figure}[t]
\includegraphics[width=\linewidth]{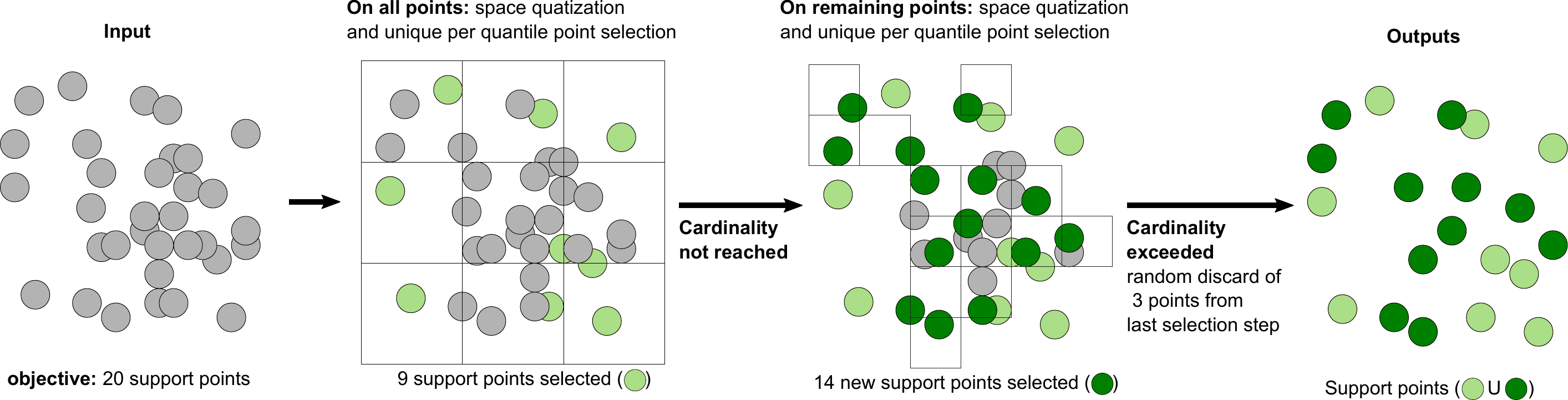}
\caption{Point sampling with space quantization.}
\label{fig:sampling}
\end{figure}

\textit{Space quantization.}
We propose an alternative approach that ensures a better sampling than~\cite{boulch2020convpoint} while being much faster than~\cite{qi2017pointnet++}.
The procedure is illustrated on Fig.~\ref{fig:sampling}.
We discretize the space using a regular voxel grid.
Each point is associated to the grid cube it falls in. In each non-empty grid cube, one point is selected. We continue with the non-selected points and a voxel size divided by two, and repeat the process until the desired number of sampled points is reached or exceeded. In the later case, some points selected at the last iteration are discarded at random to reduce the cardinality of $Q$, the set selected points.

\textit{Quantization step estimation.}
Our approach is voxel-size dependent.
On the one hand, a coarse grid leads to many iterations in the selection procedure, at the expense of computation time.
On the other hand, a fine grid reduces 
to random sampling.
Finding the optimal voxel size could be achieved using a exhaustive search (for $|Q|$ filled voxels at a single quantization step), but it is very slow.
Instead, we propose to estimate the voxel size via a rule of thumb derived by considering a simple configuration where a plane is intersecting a cube of unit length divided by a voxel grid of size $a \times a \times a$. If the plane is axis aligned, it intersects $a^2$ voxels. A sensible sampling would pick a support point in each intersected voxel, \emph{i.e.}, $|Q|=a^2$. This indicates that letting the length $v = 1/a$ of a voxel be proportional to $1/\sqrt{|Q|}$ is a reasonable choice for the voxel size. We found experimentally that choosing the diagonal length of the bounding box of the point cloud, denoted hereafter by $\rm diag$, as factor of proportionality is usually a good choice (see Section \ref{subsubsec:expe_support}). The voxel size is thus set to
\begin{equation}
\label{eq:vox_size}
    v = {\rm diag} / \sqrt{|Q|}.
\end{equation}

%% file: 04_experiments.tex
\section{Experiments}
\label{sec:expe}

In this section, we evaluate our convolutional layer on shape classification, part segmentation and semantic segmentation, reaching the state of the art regarding task metrics while being efficient regarding computation time and memory usage.

\textit{Network architectures.}
In our experiments, we use a simple yet effective residual network for classification and semantic segmentation. 
We mimic the architecture of~\cite{thomas2019kpconv}, except that ours is designed for $k$-NN convolution, i.e., we do not need to add phantom points and features to equalize the size of data tensor due to a variable number of points in radius search.
The network has an encoder-decoder structure. The encoder is composed of an alternation of residual blocks maintaining the resolution and residual blocks with down-sampling.
The decoder is a stack of fully-connected and nearest-neighbor upsampling layers.
The classification network is the encoder of the previously described network followed by a global average pooling.
For large scale semantic segmentation, we use either input modality dropout~\cite{thomas2019kpconv} or dual network fusion~\cite{boulch2020convpoint}, as indicated in tables.

\textit{Experimental setup.}
Our formulation (and code) allows a variable input size, but in order to use optimization with mini-batches, with train the networks with fixed input sizes.
As every operations of FKAConv are differentiable, all parameters are optimized via gradient descent (including the spatial gating parameters $\alpha$'s and $\beta$'s).
Finally, we use a standard cross-entropy loss.

\subsection{Benchmark results}

\begin{table}[t]
\caption{Classification and part segmentation benchmarks.}
\label{tab:bench}
\vspace{-2mm}
\centering

    \begin{minipage}{0.6\linewidth}
        \centering
        (a) ModelNet40\\~\\
        \tiny\scriptsize
        \begin{tabular}[b]{@{}l|c|@{\hspace{0.1cm}}c@{\hspace{0.2cm}}c@{}}
            \hline
            Methods & Num. & OA & AA\\
                    & points&\\
            \hline
            \textit{Mesh or voxels}\\
            Subvolume~\cite{qi2016volumetric}   &-& 89.2 & -\\
            MVCNN~\cite{su2015multi}            &-& 90.1 & -\\
            \hline
            \textit{Points} \\
            DGCNN~\cite{wang2018dynamdynamic}   &1024& 92.2 & \textbf{90.2} \\
            PointNet~\cite{qi2017pointnet}      &1024& 89.2 & 86.2 \\ 
            PointNet++~\cite{qi2017pointnet++}  &1024& 90.7 & - \\
            PointCNN~\cite{li2018pointcnn}      &1024& 92.2 & 88.1\\
            ConvPoint~\cite{boulch2020convpoint}&2048 & 92.5 & 89.6 \\
            KPConv~\cite{thomas2019kpconv}      &2048& \textbf{92.9} & - \\
            \hline
            \textit{Ours} & & \multicolumn{2}{l}{Average$\pm$std. (best run)} \\
            FKAConv  & 1024  & 92.3${\pm0.2}$ (92.5) & 89.6$\pm0.3$ (89.9)\\
                        & 2048  & 92.5${\pm0.1}$ (92.5) & 89.5$\pm0.1$ (89.7)\\
            
        \end{tabular}
    \end{minipage}
    \hfill
    \begin{minipage}{0.39\linewidth}
        \centering
        (b) ShapeNet\\~\\
        \tiny\scriptsize
        
        \begin{tabular}{l|cc}
            \hline
            Method & mcIoU & mIoU \\
            \hline
            PointNet++~\cite{qi2017pointnet++}    & 81.9 & 85.1 \\
            SubSparseCN~\cite{graham20183d}              & 83.3 & 86.0 \\
            SPLATNet~\cite{su2018splatnet}        & 83.7 & 85.4 \\
            SpiderCNN~\cite{xu2018spidercnn}      & 81.7 & 85.3 \\
            SO-Net~\cite{li2018so}                & 81.0 & 84.9 \\
            PCNN~\cite{atzmon2018point}           & 81.8 & 85.1 \\
            KCNet~\cite{shen2018mining}           & 82.2 & 83.7 \\
            SpecGCN~\cite{wang2018local}          &    - & 85.4 \\
            RSNet~\cite{huang2018recurrent}       & 81.4 & 84.9 \\
            DGCNN~\cite{wang2018dynamdynamic}     & 82.3 & 85.1 \\
            SGPN~\cite{wang2018sgpn}              & 82.8 & 85.8 \\
            PointCNN~\cite{li2018pointcnn}         & 84.6 & 86.1 \\
            ConvPoint~\cite{boulch2020convpoint}  & 83.4 & 85.8 \\
            KPConv~\cite{thomas2019kpconv}        & \textbf{85.1} & \textbf{86.4} \\
            \hline
            FKAConv (Ours)   & 84.8 & 85.7\\
        \end{tabular}
        
    \end{minipage}
\end{table}

\textit{Shape classification.}
The classification task is evaluated on ModelNet40~\cite{wu20153d}.
As the spatial pooling process is stochastic, multiple predictions with the same point cloud might lead to different outcomes.
We aggregate 16 predictions for each point cloud and select the most predicted shape (we use a similar approach for part segmentation).
On the classification task (Table~\ref{tab:bench}(a)), we present average (and best) results over five runs.
For fair comparison, we train with 1024 (resp. 2048) points.
We rank first (resp. second) among the method trained with 1024 (resp. 2048) points.
We mainly observe that increasing the number points of reduces the standard deviation of the performances.

\textit{Part segmentation.}
On ShapeNet~\cite{yi2016scalable}, the network is trained with 2048 input points and 50 outputs (one for each part). The loss and scores are computed per object category (16 object categories with 2- to 6-part labels).
The results are presented in Table~\ref{tab:bench}(b).
We rank among the best methods: top-2 or top-5 depending on the metric used, i.e., mean class intersection over union (mcIoU) or instance average intersection over union (mIoU); we are only 0.3 point mcIoU and 0.7 point mIoU behind the best method.
It is interesting to notice that we are as good as or better than several methods for which the convolution falls into our formalism, such as ConvPoint~\cite{boulch2020convpoint} or SPLATNet~\cite{su2018splatnet}.

\textit{Semantic segmentation}
Three datasets are used for semantic segmentation corresponding to three different use cases.
S3DIS~\cite{armeni20163d} is an indoor dataset acquired with an RGBD camera.
The evaluation is done using a 6-fold cross validation.
NPM3D~\cite{roynard2018paris} is an outdoor dataset acquired in four sites using a lidar-equipped car.
Finally, Semantic8~\cite{hackel2017isprs} contains 30 lidar scenes acquired statically.
NPM3D and Semantic8 are datasets with hidden test labels. Scores in the tables are reported from the official evaluation servers.

We use 8192 input points but, as
subsampling the whole scene produces a significant loss of information,
we select instead points in vertical pillars with a square footprint of 2\,m for S3DIS, and 8\,m for NPM3D and Semantic8.
The center point of the pillar is selected randomly at training time and using a sliding window at test time.
If a point is seen several times, the prediction scores are summed and the most probable class is selected afterward.

\begin{table}[t!]
\caption{Semantic segmentation benchmarks.}
\label{tab:bench_seg}
\centering
    {
    \tiny
    \begin{tabular}{l|c|c|c@{\hspace{0.1cm}}c@{\hspace{0.1cm}}c@{\hspace{0.1cm}}c@{\hspace{0.1cm}}c@{\hspace{0.1cm}}c@{\hspace{0.1cm}}c@{\hspace{0.1cm}}c@{\hspace{0.1cm}}c@{\hspace{0.1cm}}c@{\hspace{0.1cm}}c@{\hspace{0.1cm}}c@{\hspace{0.1cm}}c}
        \multicolumn{16}{c}{\scriptsize (a) S3DIS}\\
        \hline
        Method & Search & IoU & ceil. & floor & wall & beam & col. & wind. & door & chair & table & book. & sofa & board & clut. \\
        \hline
        Pointnet~\cite{qi2017pointnet}          &Knn        & 47.6 & 88.0 & 88.7 & 69.3 & 42.4 & 23.1 & 47.5 & 51.6 & 42.0 & 54.1 & 38.2 &  9.6 & 29.4 & 35.2 \\
        RSNet~\cite{huang2018recurrent}         &-          & 56.5 & 92.5 & 92.8 & 78.6 & 32.8 & 34.4 & 51.6 & 68.1 & 60.1 & 59.7 & 50.2 & 16.4 & 44.9 & 52.0 \\
        PCCN~\cite{wang2018deep}                &-          & 58.3&  92.3 & 96.2 & 75.9 & 0.27 &  6.0 & \textbf{69.5} & 63.5 & 65.6 & 66.9 & 68.9 & 47.3 & 59.1 & 46.2 \\
        SPGraph~\cite{landrieu2018large}        & Super pt. & 62.1 & 89.9 & 95.1 & 76.4 & 62.8 & 47.1 & 55.3 & 68.4 & 73.5 & 69.2 & 63.2 & 45.9 &  8.7 & 52.9 \\ 
        PointCNN~\cite{li2018pointcnn}          & Knn       & 65.4 & 94.8 & 97.3 & 75.8 & 63.3 & 51.7 & 58.4 & 57.2 & 71.6 & 69.1 & 39.1 & 61.2 & 52.2 & 58.6 \\
        PointWeb~\cite{zhao2019pointweb}        & Knn       & 66.7 & 93.5 & 94.2 & 80.8 & 52.4 & 41.3 & 64.9 & 68.1 & 71.4 & 67.1 & 50.3 & 62.7 & \textbf{62.2} & 58.5 \\
        ShellNet~\cite{zhang2019shellnet}       & Knn       & 66.8 & 90.2 & 93.6 & 79.9 & 60.4 & 44.1 & 64.9 & 52.9 & 71.6 & 84.7 & 53.8 & 64.6 & 48.6 & 59.4 \\
        ConvPoint~\cite{boulch2020convpoint} & Knn       & 68.2 & \textbf{95.0} & 97.3 & 81.7 & 47.1 & 34.6 & 63.2 & 73.2 & 75.3 & \textbf{71.8} & 64.9 & 59.2 & 57.6 & 65.0 \\
        KPConv~\cite{thomas2019kpconv}          &Radius     & \textbf{70.6} & 93.6 & 92.4 & \textbf{83.1} & \textbf{63.9} & \textbf{54.3} & 66.1 & \textbf{76.6} & 57.8 & 64.0 & \textbf{69.3} & \textbf{74.9} & 61.3 & 60.3 \\
        \hline
        FKAConv (Ours \textit{RGB only})            & Knn       & 64.9 & 94.0 &	97.8 &	80.5 &	38.5 &	48.5 &	49.8 &	68.0 &	79.4 &	70.7 &	48.4 &	43.7 &    62.9 &	61.4 \\
        FKAConv (Ours \textit{RGB drop.})           & Knn       & 66.6 & 94.4	& 97.8	& 81.5	& 38.7	& 43.3	& 56.4	& 71.6	& 80.2	& 71.8	& 63.5	& 54.1	& 50.6	& 62.5 \\
        \hline
        FKAConv (Ours \textit{fusion})              & Knn       & 68.4 & 94.5 & \textbf{98.0} & 82.9 & 41.0 & 46.0 & 57.8 & 74.1 & \textbf{77.7} & 71.7 & 65.0 & 60.3 & 55.0 & \textbf{65.5} \\
        Rank                                    &           & 2    & 3    & 1    & 2    & 7    & 4    & 6    & 2    & 1    & 2    & 3    & 4    & 5    & 1 \\
      \end{tabular}
    }
    
    \vspace{0.2cm}
    {
    \tiny
    \begin{tabular}{@{}l|c|ccccccccc@{}}
    \multicolumn{11}{c}{\scriptsize (b) NPM3D}\\
        \hline
        Method & Av.IoU & Ground & Building & Pole & Bollard & Trash can & Barrier & Pedestrian & Car & Natural \\ \hline
        RF MSSF~\cite{thomas2018semantic}	        & 56.3	& 99.3	& 88.6	& 47.8	& 67.3	& 2.3	& 27.1	& 20.6	& 74.8	& 78.8 \\
        MS3 DVS~\cite{roynard2018classification}	        & 66.9	& 99.0	& 94.8	& 52.4	& 38.1	& 36.0	& 49.3	& 52.6	& 91.3	& 88.6 \\
        HDGCN~\cite{liang2019hierarchical}	        & 68.3	& 99.4	& 93.0	& 67.7	& 75.7	& 25.7	& 44.7	& 37.1	& 81.9	& 89.6 \\
        ConvPoint~\cite{boulch2020convpoint}	    & 75.9	& 99.5	& 95.1	& 71.6	& 88.7	& 46.7	& 52.9	& 53.5	& 89.4	& 85.4 \\
        KPConv~\cite{thomas2019kpconv}	        & 82.0	& 99.5	& 94.0	& 71.3	& 83.1	& \textbf{78.7}	& 47.7	& \textbf{78.2}	& 94.4	& 91.4 \\
        \hline
        FKAConv	(ours \textit{fusion})      & \textbf{82.7}	& \textbf{99.6}	& \textbf{98.1}	& \textbf{77.2}	& \textbf{91.1}	& 64.7	& \textbf{66.5}	& 58.1	& \textbf{95.6}	& \textbf{93.9} \\
        Rank            & 1     & 1     & 1     & 1     & 1     & 2     & 1     & 2     & 1     & 1     \\
        \multicolumn{10}{l}{\textit{Note: We report here only the published methods at the time of writing.}}
    \end{tabular}
    }
    
    \vspace{0.2cm}
    {
    \tiny
    \begin{tabular}{l|c@{~}c|c@{~}c@{~}c@{~}c@{~}c@{~}c@{~}c@{~}c}
        \multicolumn{11}{c}{\scriptsize (c) Semantic3D}\\ \hline
        Method          & Av. & OA  & Man   & Nat.   & High  & Low   & Build. & Hard  & Art. & Cars\\
                        & IoU &     & made  &           & veg.  & veg.  &           & scape & \\
        \hline
        TML-PC~\cite{montoya2014mind}	        & 39.1	& 74.5	& 80.4	& 66.1	& 42.3	& 41.2	& 64.7	& 12.4	& 0.	& 5.8 \\
        TMLC-MS~\cite{hackel2016fast}           & 49.4	& 85.0	& 91.1	& 69.5	& 32.8	& 21.6	& 87.6	& 25.9	& 11.3	& 55.3 \\
        PointNet++~\cite{qi2017pointnet++}      & 63.1	& 85.7	& 81.9	& 78.1	& 64.3	& 51.7	& 75.9	& 36.4	& 43.7	& 72.6 \\
        EdgeConv~\cite{contreras2019edge}       & 64.4	& 89.6	& 91.1	& 69.5	& 65.0	& 56.0	& 89.7	& 30.0	& 43..8	& 69.7 \\
        SnapNet~\cite{boulch2017snapnet}        & 67.4	& 91.0	& 89.6	& 79.5	& 74.8	& 56.1	& 90.9	& 36.5	& 34.3	& 77.2 \\
        PointGCR~\cite{ma2020global}                         & 69.5	& 92.1	& 93.8	& 80.0	& 64.4	& 66.4	& 93.2	& 39.2	& 34.3	& 85.3 \\
        FPCR~\cite{truong2019fast}	                            & 72.0	& 90.6  & 86.4	& 70.3	& 69.5  & 68.0	& 96.9	& 43.4	& 52.3	& 89.5 \\
        SPGraph~\cite{landrieu2018large}        & 76.2	& 92.9	& 91.5	& 75.6	& \textbf{78.3}	& 71.7	& 94.4	& \textbf{56.8}	& 52.9	& 88.4 \\
        ConvPoint~\cite{boulch2020convpoint} & \textbf{76.5}	& 93.4	& 92.1	& 80.6	& 76.0	& \textbf{71.9}	& \textbf{95.6}	& 47.3	& \textbf{61.1}	& 87.7 \\
        \hline
        FKAConv* (ours fusion)                   & 74.6	& \textbf{94.1}	& \textbf{94.7}	& \textbf{85.2}	& 77.4	& 70.4	& 94.0	& 52.9	& 29.4	& \textbf{92.6} \\
        Rank                                & 3	    & 1	    & 1	    & 1	    & 2	    & 3	    & 5	    & 2	    & 9	    & 1 \\
        \multicolumn{10}{l}{\textit{Note: We report here only the published methods at the time of writing.}}\\
        \multicolumn{10}{l}{\textit{*In the official benchmark, the entry corresponding to our method is called LightConvPoint,}}\\
        \multicolumn{10}{l}{\textit{~~which refers to the framework used for our implementation.}}
        
    \end{tabular}
    }
\end{table}

\begin{figure}
    \centering
    \newcommand{\tlap}[1]{\vbox to 0pt{\vss\hbox{#1}}}
    
    \begin{tabular}{ccccc}
    \includegraphics[width=0.19\linewidth]{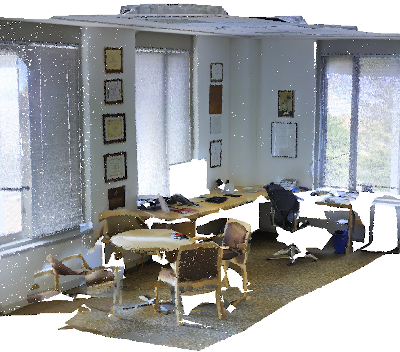}&
    \includegraphics[width=0.19\linewidth]{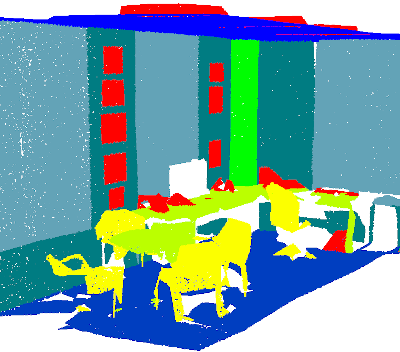}&
    \includegraphics[width=0.19\linewidth]{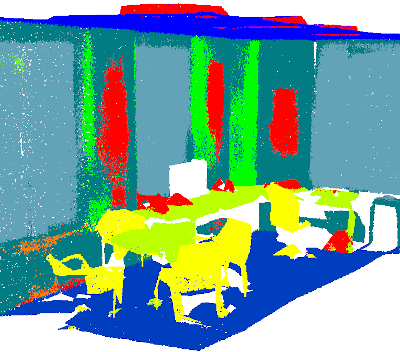}&
    \includegraphics[width=0.19\linewidth]{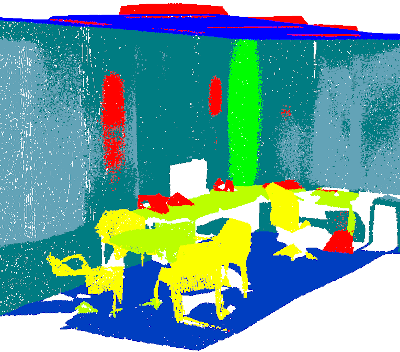}&
    \includegraphics[width=0.19\linewidth]{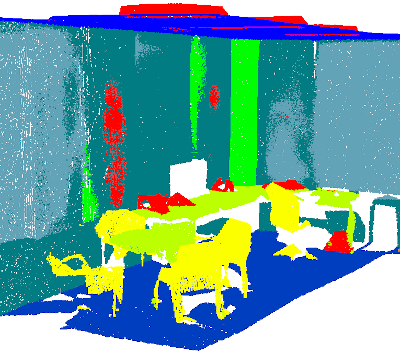}\\
    \textit{\llap{In}put:\,3D\,\,w/RG\rlap{B}} & \textit{~Groung truth} & \textit{PointWeb}\cite{zhao2019pointweb} & \textit{ConvPoint}\cite{boulch2020convpoint} & \textit{\llap{F}KAConv (ours\rlap)}
    \end{tabular}
    
    \vspace{0.2cm}
    
    \begin{tabular}{cccc}
        \includegraphics[width=0.24\linewidth]{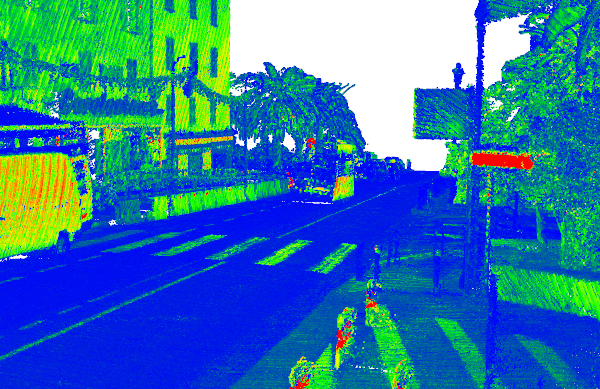}&  
        \includegraphics[width=0.24\linewidth]{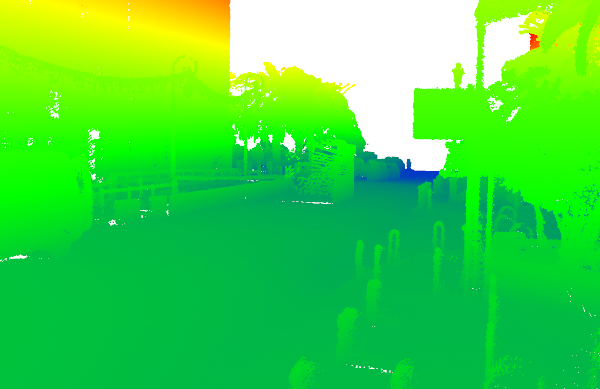}&  
        \includegraphics[width=0.24\linewidth]{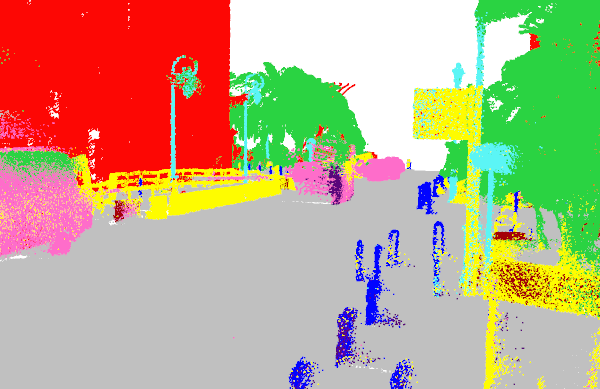}&  
        \includegraphics[width=0.24\linewidth]{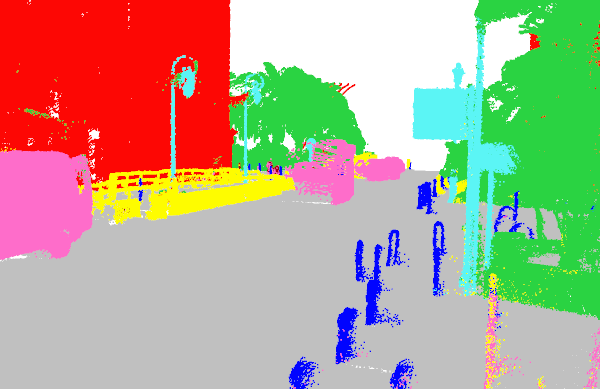} \\
        \textit{\llap{In}:\,3D\,w/laser\,intens\rlap{ity}} & \textit{Elevation view} & \textit{ConvPoint~\cite{boulch2020convpoint}} & \textit{FKAConv (ours)}
    \end{tabular}
    
    \vspace{0.2cm}
    
    \begin{tabular}{cccc}
        \includegraphics[width=0.24\linewidth]{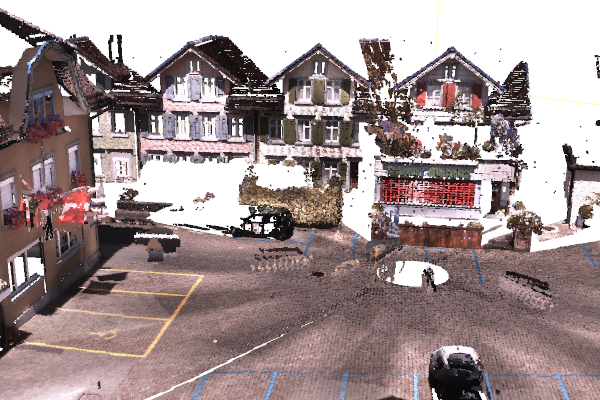} &  
        \includegraphics[width=0.24\linewidth]{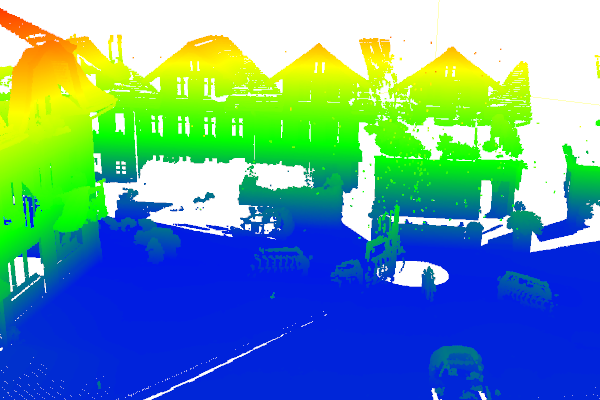} &
        \includegraphics[width=0.24\linewidth]{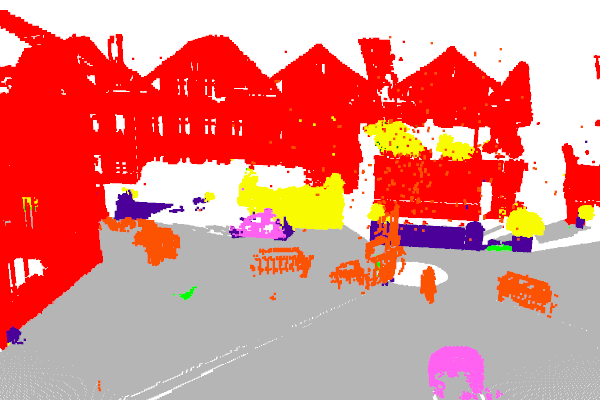} &
        \includegraphics[width=0.24\linewidth]{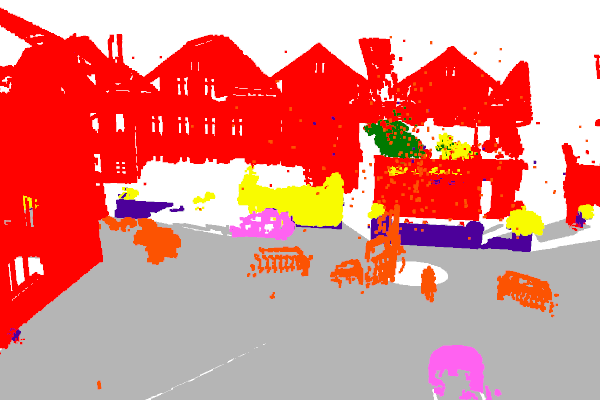} \\
        \textit{Input: 3D with RGB} & \textit{Elevation view} & \textit{ConvPoint~\cite{boulch2020convpoint}} & \textit{FKAConv (ours)}
    \end{tabular}
    
    \caption{Visual results of semantic segmentation: S3DIS (1\tlap{\textsuperscript{st}} row), NPM3D (2\tlap{\textsuperscript{nd}} row) and Semantic3D (3\tlap{\textsuperscript{rd}} row). Ground truth of test data publicly unavailable for last two.}
    \label{fig:sem_visu}
\end{figure}

The results are presented in Fig.~\ref{fig:sem_visu} and Table~\ref{tab:bench_seg}.
We use S3DIS (Table~\ref{tab:bench_seg}(a)) to study the impact of the training strategy.
As underlined in~\cite{thomas2019kpconv,boulch2020convpoint}, direct learning with colored points yields a model relying too much on color information, at the expense of geometric information.
We train three models.
The first is the baseline model trained with color information, the second uses color dropout as in~\cite{thomas2019kpconv}, and the third is a dual model with a fusion module~\cite{boulch2020convpoint}.
We observe that fusion gives the best results.
In practice, the model trained with modality dropout tends to select one of the two modalities, either color or geometry, depending on what modality gives the best results.
On the contrary, the fusion technique uses two networks each trained with a different modality, resulting in a lot larger network, but ensuring that the information of both modalities is taken into account.

Our network is second on S3DIS, first on NPM3D and third on Semantic8.
On S3DIS, it is the best approach for 3 out of 13 categories and it performs well on the remaining ones.
We are only outperformed by KPConv, which is based on radius search.
On NPM3D, we reach an average intersection over union (av. IoU) of 82.7, which is 0.7 point above the second best method.
On Semantic8, we place third according to average IoU, and first on overall accuracy among the published and arXiv methods. We obtain the best scores in 3 out 8 categories (the top-3 for 6 categories out of 8). More interestingly, we exceed the scores of ConvPoint~\cite{boulch2020convpoint} on 5 categories.
The only downside is the very low score on the category of artefacts.
One possible explanation could be that the architecture used in this paper (the residual network) is not suitable to learn a reject class (the artefact class is mainly all the points that do not belong to the 7 other classes, i.e., pedestrians but also scanning outliers).
It is future work to train the ConvPoint network with our convolution layer to support this hypothesis.

\subsection{Support point sampling: discretization parameter.}
\label{subsubsec:expe_support}

\begin{figure}[t]
    \centering
    \includegraphics[width=0.5\linewidth]{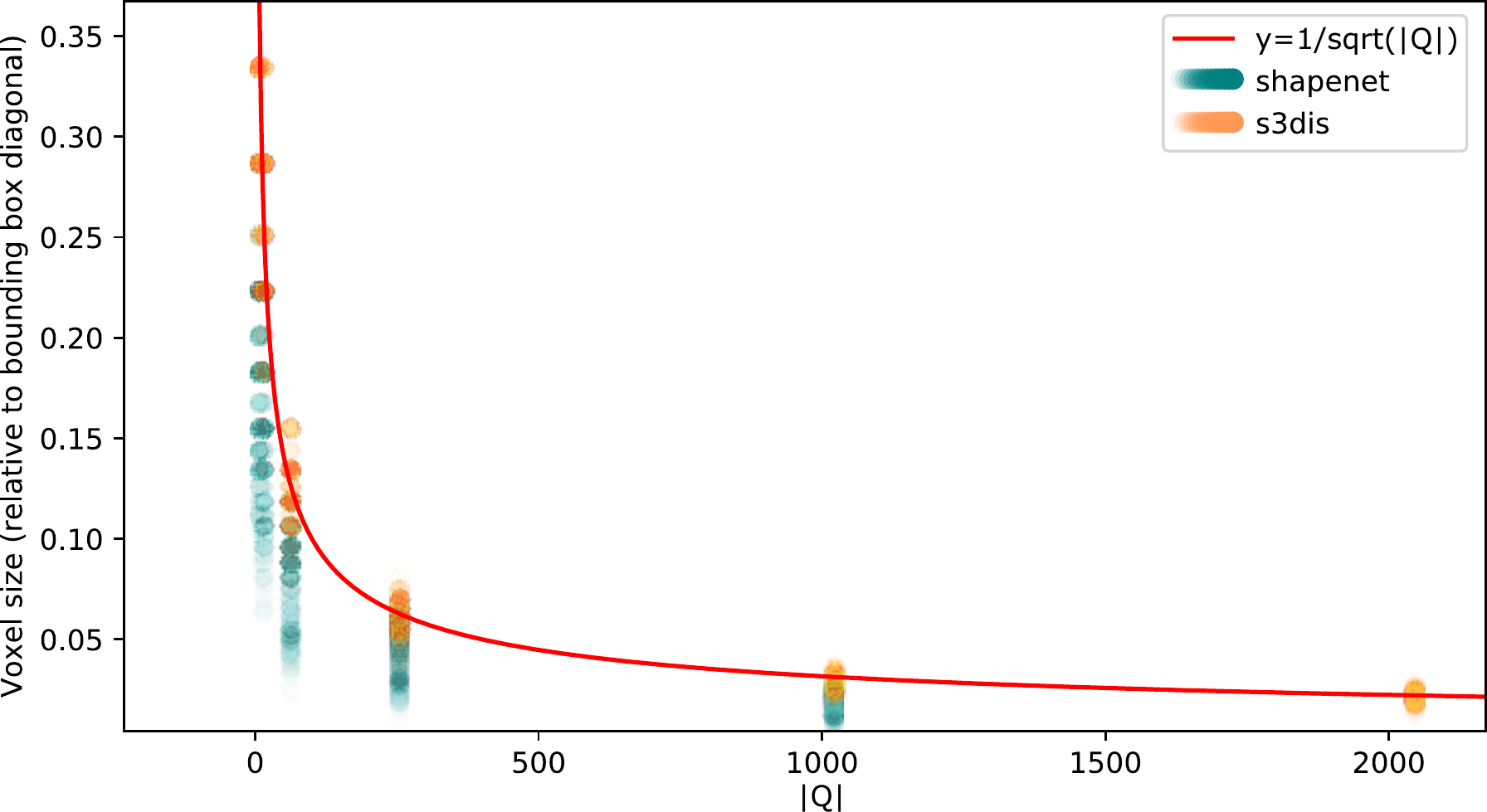}
    \caption{Empirical validation of voxel size estimation: ShapeNet (blue), S3DIS (orange). Each dot is the empirical optimal voxel size obtained by dichotomic search. The red line is the voxel size defined as the inverse square root of the number of support points.}
    \label{fig:voxsize}
\end{figure}

The rule of thumb in Equation~\eqref{eq:vox_size} was derived in a simplistic case: a point cloud sampled from an axis aligned plane crossing a regular voxel grid.
In practice, planar surfaces are very common, particularly in semantic segmentation (walls, floors, etc.), but are not a good model for most of the object of the scenes (chairs, cars, vegetation, \emph{etc.}).
To validate Eq.\,\eqref{eq:vox_size}, we compute the optimal quantization parameter (i.e., the parameter with the largest value leading to the desired number of support points in a single quantization) computed using a dichotomic search on the parameter space and compare it to the derived expression. 
Figure~\ref{fig:voxsize} presents the results of the experiment.
For each point cloud, the optimal voxel size is represented by a semi-transparent disk (blue for ShapeNet, orange for S3DIS) and can be compared to the derived expression (red curve).
In our setting, a curve under the colored disks is not desired; it is an over-quantization.
We prefer a curve above these disks, possibly leading to extra iterations, but not affecting performance.
We observe that Equation~\eqref{eq:vox_size} provides a good estimate of the voxel size, especially for S3DIS which is a dataset containing a lot of planes.
For ShapeNet, we observe a higher variance, due to the great variability of shapes.
Because of numerous objects that cannot be modeled well by planes in ShapeNet, we slightly overestimate the voxel size, leading only to one spurious iteration, which only slightly slows down the operation.

\subsection{Support point sampling: computation times.}

To assess our sampling approach, we run two experiments.
First, in Table~\ref{tab:timing_sampling}(a), we compare the sampling time as a function of the size of the input point cloud.
The number of support points is half the input point cloud size, and the number of neighbors is $16$. The scores are averaged over 5000 random points clouds sampled in a cube.
We also report the ShapeNet scores to relate the performance and the computation times.
We compare our sampling strategy with farthest point sampling~\cite{qi2017pointnet++}, with iterative neighborhood rejection~\cite{boulch2020convpoint} and with a random baseline.
As farthest point sampling~\cite{qi2017pointnet++} is the reference of several state-of-the-art methods, we give the gain relatively to this method in percentage.
Our quantized sampling is almost as fast as random sampling and much more efficient than farthest point sampling.
In fact, our sampling has almost a linear complexity, compared to farthest point sampling, that has a quadratic complexity.

\begin{table}[t]
\caption{Computation time and memory consumption.}
\label{tab:timing_sampling}
\centering

\begin{minipage}{0.49\linewidth}
(a) Computation times for different\\
sampling strategies.
~\\~\\
{
\tiny
\begin{tabular}{l|r|r|r|c}
Method & \multicolumn{3}{c|}{Sampling time (ms)} & ShapeNet\\
       & 1k pts & 5k pts    &   10k pts & (mIoU)\\
\hline
Random     & 1.66 & 8.6 & 18.6 & 84.4\\
(baseline) & ({\color{ForestGreen}-60\%}) & ({\color{ForestGreen}-89\%}) &({\color{ForestGreen}-94\%}) \\
\hline
ConvPoint~\cite{boulch2020convpoint} & 2.60    & 25.4     & 88.2   & 84.6 \\
& ({\color{ForestGreen}-37\%}) & ({\color{ForestGreen}-68\%}) & ({\color{ForestGreen}-71\%}) \\
Farthest~\cite{qi2017pointnet++}        & 4.12    & 79.8    & 310.2   & 84.7\\
 &(-) &(-) &(-)\\
\hline
FKAConv sampling                            &  1.93   & 10.3   & 20.4 & 84.6\\
 &({\color{ForestGreen}-53\%}) & ({\color{ForestGreen}-87\%})   & ({\color{ForestGreen}-93\%})\\
\multicolumn{5}{l}{~~~~(time for $n$ inputs points, $n/2$ support points,}\\
\multicolumn{5}{l}{~~~~~$16$ neighbors, averaged over $5000$ iterations).}
\end{tabular}
}

\end{minipage}
\hfill
\begin{minipage}{0.49\linewidth}

(b) Time and memory consumption for a segmentation network, with 8192 points.
~\\~\\
{
\tiny
\begin{tabular}{l|rr|rr}
Convolution & \multicolumn{2}{c|}{Training} & \multicolumn{2}{c}{Test}\\
Layer       & Time  & Memory& Time  & Memory\\
            & (ms)  & (GB)  & (ms)  & (GB)  \\
\hline
ConvPoint~\cite{boulch2020convpoint} & 85.7 & 10.1   & 65    & 2.9   \\
ConvPoint*                              & \textbf{12.2} & \textbf{4.3}    & 4.29  & 1.6  \\
PointCNN*~\cite{li2018pointcnn}         & 33.6 & 3.5    & 6.23  & 1.7  \\
PCCN*~\cite{wang2018deep}               & 31.1 & 4.9    & 10.2  & 2.3  \\
PCCN** (bs4)                            & 64.2 & 6.4    & 19.7  & 2.6  \\
\hline
FKAConv (Ours)                              & 19.1 & 5.6   & \textbf{4.9}  & \textbf{1.4}  \\
\multicolumn{5}{l}{*: reimplemented in our framework.}\\
\multicolumn{5}{l}{**: original formulation without separation trick,}\\
\multicolumn{5}{l}{~~~~~differs from code used for experiments in~\cite{wang2018deep}.}
\end{tabular}
}
\end{minipage}
\end{table}

\subsection{Inference time and memory consumption}

We present in Table~\ref{tab:timing_sampling}(b) the performance of our convolution layer and compare it to other convolutional layers.
All computation times and memory usage are given for the segmentation network architecture and for one point cloud. The measures were done with $8192$ points in each point cloud and a batch size of $16$ (except for PCCN** for which the batch size is reduce to $4$ to fit in the 11 GB GPU memory).
Computational times are given per point cloud in milliseconds, and memory usage is reported in gigabytes.

We observe that our computation times at inference are very similar to those of ConvPoint~\cite{boulch2020convpoint}, which is expected as it falls into our same general formulation.
The same would probably be observed for a $k$-NN version of the KPConv~\cite{thomas2019kpconv}.
Then, we remark that PointCNN~\cite{li2018pointcnn} and PCCN~\cite{wang2018deep} are up to twice slower for inference.
PCCN uses the separable kernel trick to improve memory performance (cf.\ Section~\ref{sec:kernelsep}).
In this form, it is similar to $N_{\text{out}}$ ($N_{\text{out}}$ being the number of filters of the layer) parallel instances of our layer with one kernel element, i.e., it is equivalent to estimating a different $\mathbf{A}$ for each $f \in \{1,\dots,F\}$.
We also report in Table~\ref{tab:timing_sampling}(b) the performance for PCCN**, which is the the purely continuous convolution described in PCCN~\cite{wang2018deep}, but without the separable kernel trick.

\subsection{Filter visualization}

\begin{figure}[t]
\centering

\begin{tabular}{c@{}c@{}c@{}c@{~}|c@{}@{~}c@{}c@{}c@{}c}
    \multicolumn{4}{c}{Low-level features} & \multicolumn{5}{c}{High-level features}\\
    ~\\
    \includegraphics[width=0.10\linewidth]{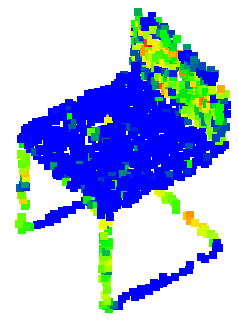}&
    \includegraphics[width=0.10\linewidth]{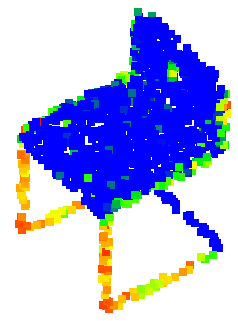}&
    \includegraphics[width=0.10\linewidth]{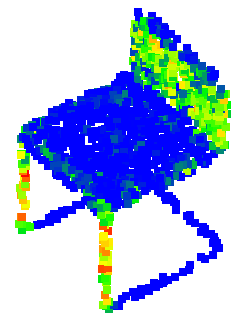}&
    \includegraphics[width=0.10\linewidth]{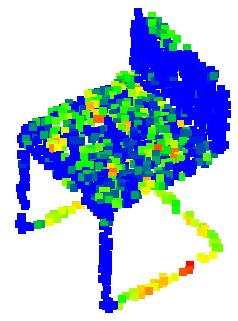}&
     \includegraphics[width=0.10\linewidth]{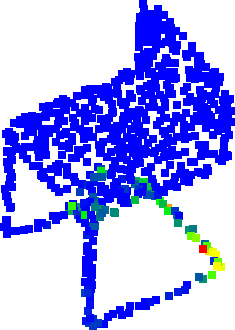}&
     \includegraphics[width=0.10\linewidth]{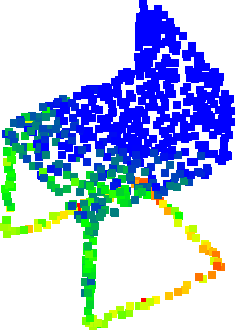}&
     \includegraphics[width=0.10\linewidth]{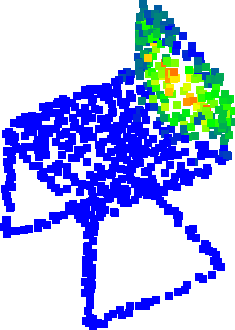}&
     \includegraphics[width=0.10\linewidth]{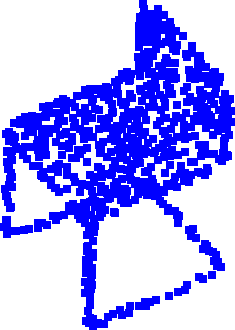}&
     \includegraphics[width=0.10\linewidth]{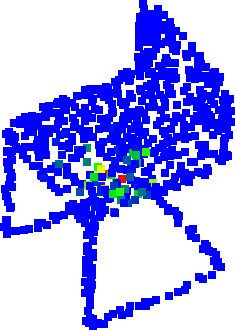}\\
     \includegraphics[width=0.10\linewidth]{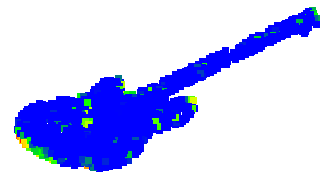}&
     \includegraphics[width=0.10\linewidth]{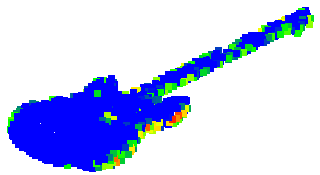}&
     \includegraphics[width=0.10\linewidth]{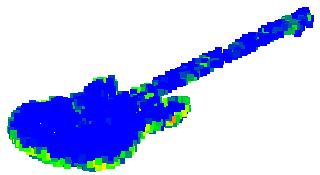}&
     \includegraphics[width=0.10\linewidth]{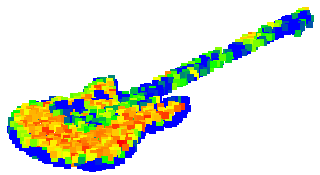}&
     \includegraphics[width=0.10\linewidth]{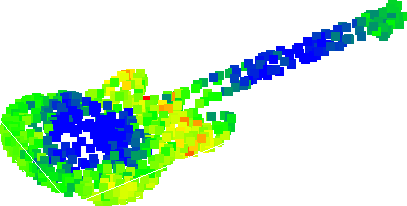}&
     \includegraphics[width=0.10\linewidth]{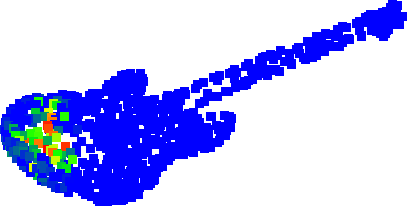}&
     \includegraphics[width=0.10\linewidth]{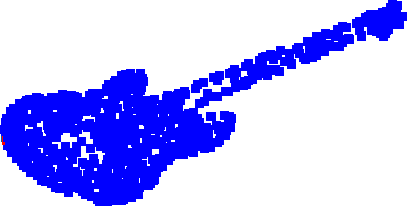}&
     \includegraphics[width=0.10\linewidth]{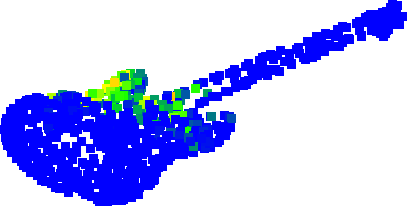}&
     \includegraphics[width=0.10\linewidth]{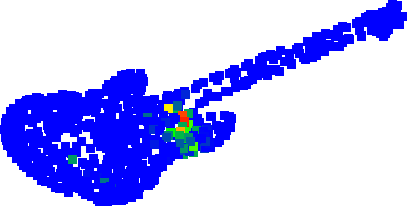}\\
     \includegraphics[width=0.10\linewidth]{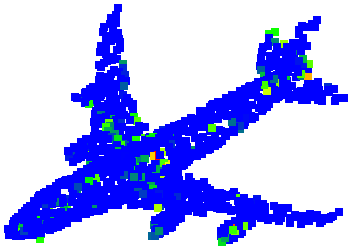}&
     \includegraphics[width=0.10\linewidth]{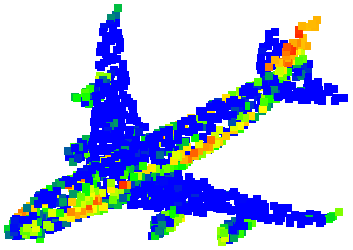}&
     \includegraphics[width=0.10\linewidth]{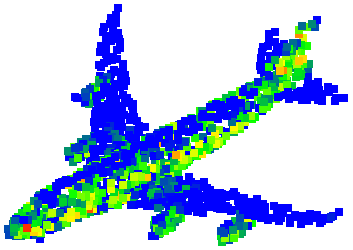}&
     \includegraphics[width=0.10\linewidth]{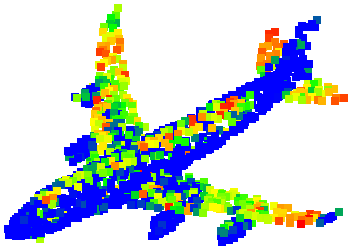}&
     \includegraphics[width=0.10\linewidth]{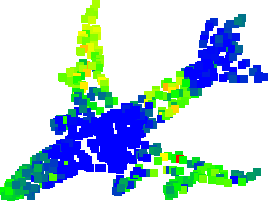}&
     \includegraphics[width=0.10\linewidth]{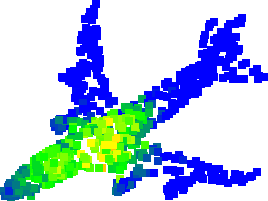}&
     \includegraphics[width=0.10\linewidth]{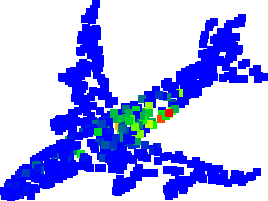}&
     \includegraphics[width=0.10\linewidth]{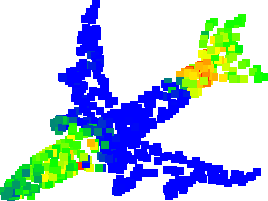}&
     \includegraphics[width=0.10\linewidth]{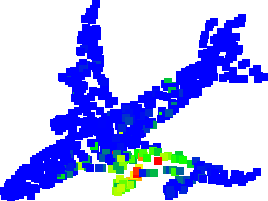}
\end{tabular}

    \caption{FKAConv filter response for different input shapes on ModelNet40. Low-level features are extracted from the first layer (4 filters), and high-level features from the fourth layer (5 filters). The colormap represents the filter response for the shape, from blue (low response) to red (high response).}
    \label{fig:filter}
\end{figure}

Our method FKAConv was derived from the discrete convolution on regular grids. The behavior of our 3D filters should thus be comparable to their 2D counterparts.
In Figure~\ref{fig:filter}, we present the outputs of early and deep filters for the classification network on ModelNet40.
For easier visualization, the features at coarse scales (high level / deep features) have been upsampled at the full point-cloud resolution.
We notice that early layers produce features based on surface orientation. This is consistent with the small receptive field of early layers, that yields fine-scale features.
On the contrary, deep layers produces shape-related features detecting objects parts, such as people heads or airplane bodies.

%% file: 05_conclusion.tex
\section{Conclusion}

We presented a formulation for convolution on point clouds that unifies a range of existing convolutional layers and suggests a new point convolution approach. The core of the method is the estimation of an alignment matrix between the input points and the kernel.
We also introduced an alternative point sampling strategy to farthest point sampling by using a progressive voxelization of the input space. While being almost as efficient as farthest point sampling, it is nearly as fast as random sampling. With these conceptually simple and easy to implement ideas, we obtained  state-of-the-art results on several classification and semantic segmentation benchmarks among methods based on $k$-NN search, while being among the fastest and most memory-efficient approaches.

%% file: 06_supplementary.tex
\appendix

We present complementary information about the FKAConv paper.
In Section~\ref{supp:sec:arch}, we detail the network architectures used for classification and semantic segmentation.
In Section~\ref{supp:sec:datasets}, we briefly describe the datasets used for experiments.
In Section~\ref{supp:sec:knn_rad}, we discuss the use of a learned normalization of the support point neighborhoods.
Finally, in Section~\ref{supp:sec:qualitative}, we provide more qualitative results on the semantic segmentation test datasets.

\section{Network details}
\label{supp:sec:arch}

As mentioned in Section~5.1 of the paper, the networks we use in our experiments are based on the residual network architecture in KPConv~\cite{thomas2019kpconv}, for which we replace the convolution by FKAConv and the neighborhood construction by our point sampling with space quantization.
We detail here the main components.

\begin{figure}[p]
    \begin{tabular}{c}
        \includegraphics[width=0.98\linewidth]{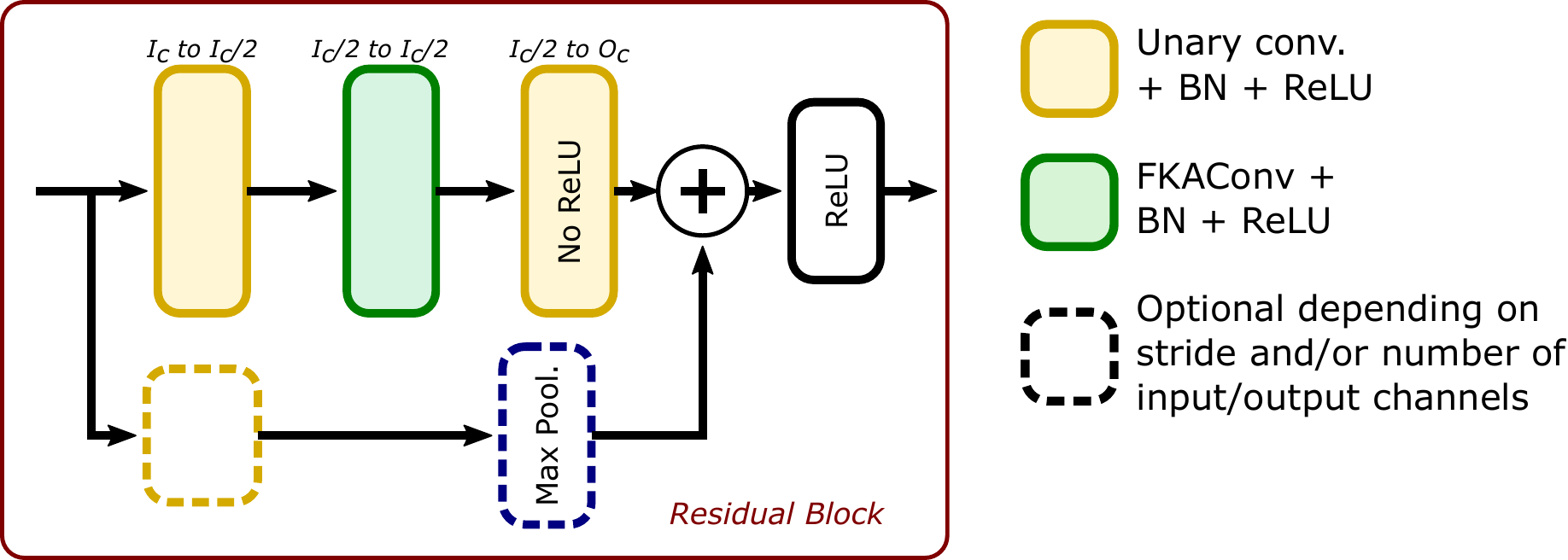}\vspace{2mm}\\ 
        (a) Residual block. $I_c$ (resp. $O_c$) is the number of input (resp. output) channels.\\~\\~\\
        \includegraphics[width=\linewidth]{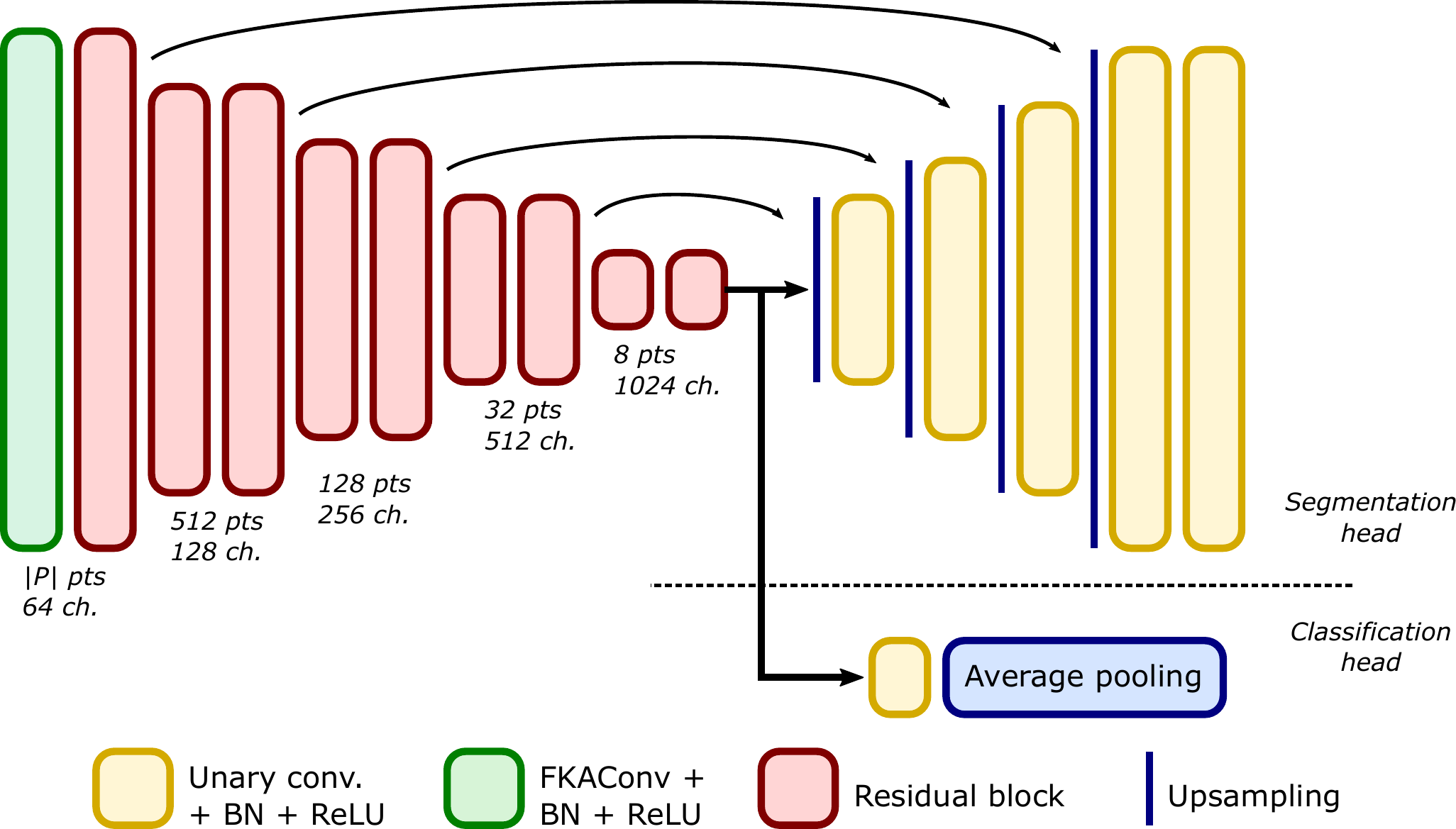}\vspace{2mm}\\
        (b) Classification and segmentation networks.\\~\\
        \includegraphics[width=0.9\linewidth]{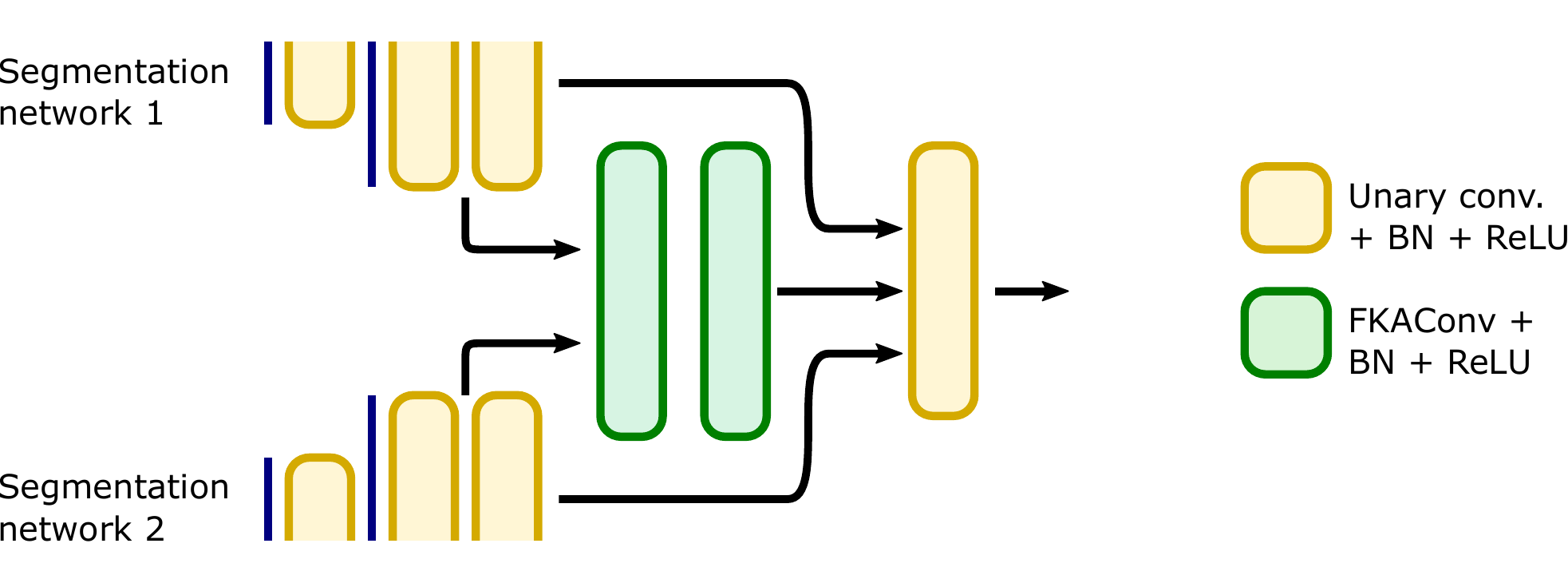}\\
        (c) Fusion module.
    \end{tabular}
    \centering
    
    \caption{Network architectures used for semantic segmentation and classification.}
    \label{supp:fig:net_kp}
\end{figure}

\subsubsection{Residual block.}
The residual block in Fig.\,\ref{supp:fig:net_kp}\,(a) is the main module of our networks.
This block is made of an FKAConv layer placed between two linear layers.
The residual connection has one \textit{optional} linear layer and one \textit{optional} max-pooling layer.
The \textit{optional} linear layer is used only when the number of input channels is different from the number of output channels, and the \textit{optional} max-pooling layer is used if the cardinality of the support points is different from the cardinality of the input points. 

\subsubsection{Classification and segmentation networks.}

The two networks for these tasks are presented in Fig.\,\ref{supp:fig:net_kp}\,(b).
They share the same encoder structure, i.e., a FKAConv layer and 9 residual modules with a progressive reduction of the point cloud size.
\begin{itemize}
    \item The classification network has an extra point-wise fully-connected layer (or unary convolutional) with its output dimension equal to the number of classes.
The final prediction is done by averaging the scores of the 8 final support points.
    \item The segmentation network has an encoder-decoder structure.
    The decoder is a stack of 5 unary layers with, nearest-neighbor up-sampling between each layer.
    We use skip connections from the encoder to the decoder: the target points for up-sampling are the support points at the corresponding scale in the encoder, and the features from the encoder and the decoder are concatenated at each scale.
\end{itemize}

\subsubsection{Fusion network.}

The fusion module presented in Fig.\,\ref{supp:fig:net_kp}\,(c) is identical to the module from the official repository of ConvPoint~\cite{boulch2020convpoint}, except that it uses our proposed convolution.
It is made of 3 layers: 2 FKAConv layers and 1 unary layer.
The features from the penultimate layer of both segmentation networks are concatenated and given as input to the first FKAConv layer.
The output of the second FKAConv layer is then concatenated with the predictions of the two segmentation networks and given to the unary layer.

\subsubsection{Parameters of the convolution.}
In order to keep the setup simple, we use the same parameters for all FKAConv layers.
The neighborhood size of the support points is fixed to 16 and we use 16 kernels.
In the encoder, each odd residual block but the first reduces the number of support points from $|$P$|$ to 512, 128, 32, and finally 8.

\section{Datasets}
\label{supp:sec:datasets}

We evaluate our convolution on three different tasks: object classification, part segmentation and semantic segmentation.

\subsubsection{Classification} is evaluated on ModelNet40~\cite{wu20153d}.
It contains 12,311 point clouds sampled from CAD models of 40 different categories.
9843 shapes are used for training, 2468 for testing.

\subsubsection{Part segmentation} is done on Shapenet~\cite{yi2016scalable}.
It is composed of 16 object category, each category being annotated with 2 to 6 part labels.
As a pre-processing, all models are first normalized to the unit sphere.
In our implementation, the network has 50 outputs (one for each part) and the loss and scores are computed per object category.

\subsubsection{Semantic segmentation} is evaluated on 3 different indoor and outdoor datasets: S3DIS~\cite{armeni20163d}, NPM3D~\cite{roynard2018paris} and Semantic8~\cite{hackel2017isprs}.
\begin{itemize}
    \item  S3DISS~\cite{armeni20163d} is a subset of the 2D-3D-S dataset for semantic segmentation of building interiors.
    The data are acquired over 6 building floors with an RGBD camera.
    Each points is annotated with one of 13 labels: 12 semantic labels (floors, tables, chairs, etc.) and 1 label for a ``clutter'' class, mostly including office supplies.
    The evaluation is done using a 6-fold cross validation.
    
    \item NPM3D~\cite{roynard2018paris} is a lidar dataset for large-scale outdoor semantic segmentation.
    Points were acquired in 4 sites using a car equipped with lidar.
    10 classes of urban entities are labeled, such as impervious surface, poll or pedestrian.
    
    \item Semantic8~\cite{hackel2017isprs} is the main dataset in the Semantic3D benchmark suite.
    It contains 30 lidar scenes, 15 for training and 15 for testing.
    Over 4 billion points are labeled with 8 classes such as building, vegetation and car, and a challenging class for scanning artifacts.
    The test set is particularly difficult as it covers several diverse scenes such as city streets, villages or old castles.
\end{itemize}

\section{$K$-nearest neighbors search with learned neighborhood normalization}
\label{supp:sec:knn_rad}

As described in Section~4 of the paper, we normalize the size of point neighborhoods.  After recalling the principle of this normalization, we analyze here its effect empirically, for different datasets and for different layers in our networks. 

To normalize the neighborhoods, we estimate an average neighborhood radius $r_t$ using an exponential average of the actual neighborhood radius $\hat{r}_t$ seen at training time in the successive batches indexed by $t$ (see Equation~\eqref{eq:radius} below where $m$ is the momentum parameter). The point coordinates $(\mathbf{p}_i)_{1\leq i\leq k}$ of the $k$ nearest neighbors of a support point $\mathbf{q}$ are then centered and normalized as specified in Equation~\eqref{eq:localcoord}, yielding normalized points $(\hat{\mathbf{p}}_i)_{1\leq i\leq k}$:
\begin{eqnarray}
r_{t} &=& \hat{r}_t * m + r_{t-1} * (1-m), \label{eq:radius}\\
\hat{\mathbf{p}}_i &=& (\mathbf{p}_i-\mathbf{q}) / r_{t}. \label{eq:localcoord}
\end{eqnarray}
We also proposed a gating mechanism to reduce, if needed, the negative effect of faraway points.
Given the centered and normalized points $(\hat{\mathbf{p}}_i)_i$ computed above, the spatial gate weight $\mathbf{s} = (s_i)_i$ satisfies:
\begin{equation}
    s_i = \sigma(\beta - \alpha ||\hat{\mathbf{p}}_i||_2),
\end{equation}
where $\sigma(\cdot)$ is the sigmoid function, and $\alpha$, $\beta$ are parameters to learn.


\begin{figure}[p]
    \centering
    \begin{tabular}{c@{~~~~}c}
        \multicolumn{2}{c}{\includegraphics[width=0.48\linewidth]{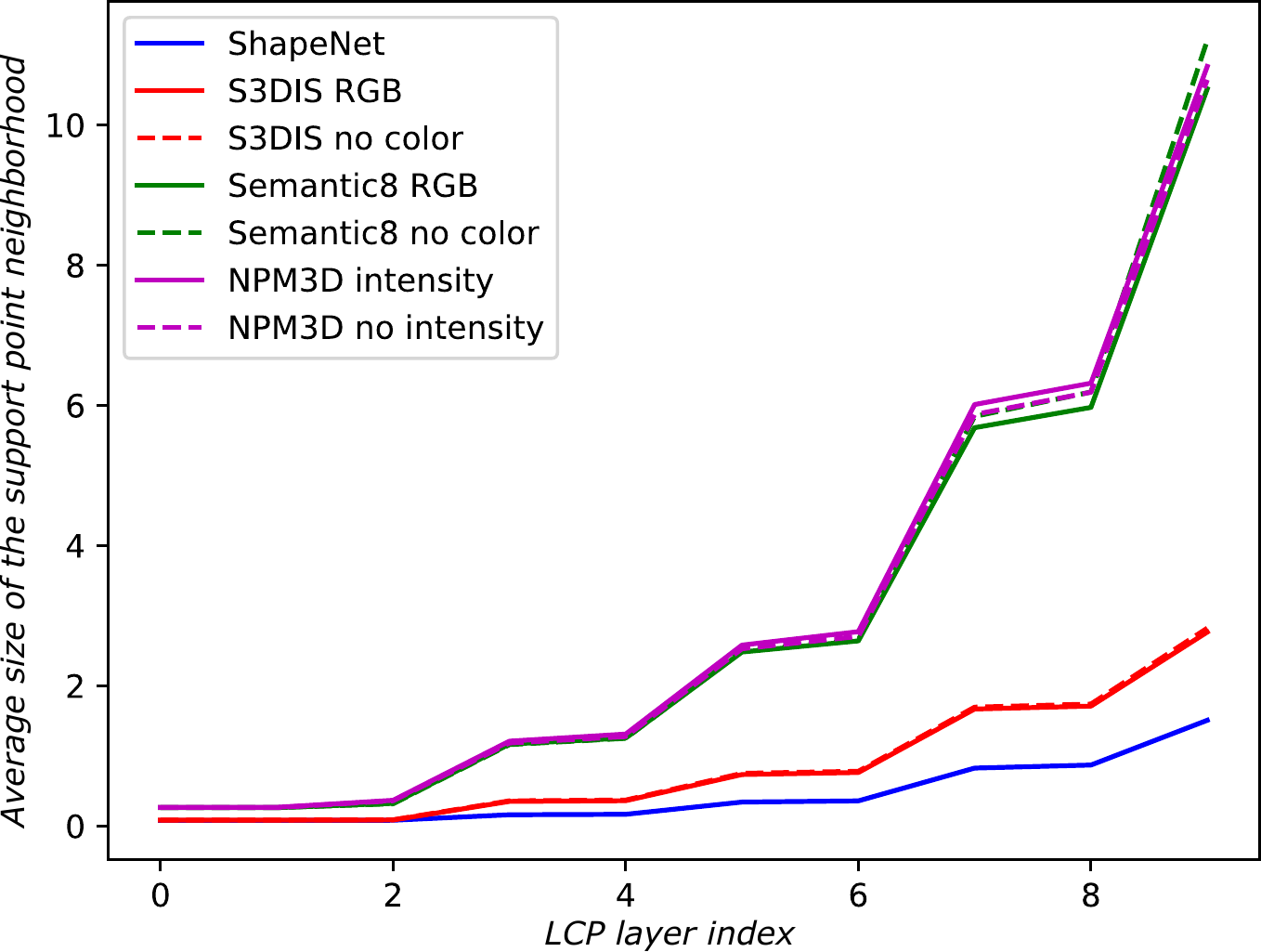}}\\
        \multicolumn{2}{c}{(a) Average neighborhood radius at each layer.} \\
        ~\\~\\
        \includegraphics[width=0.48\linewidth]{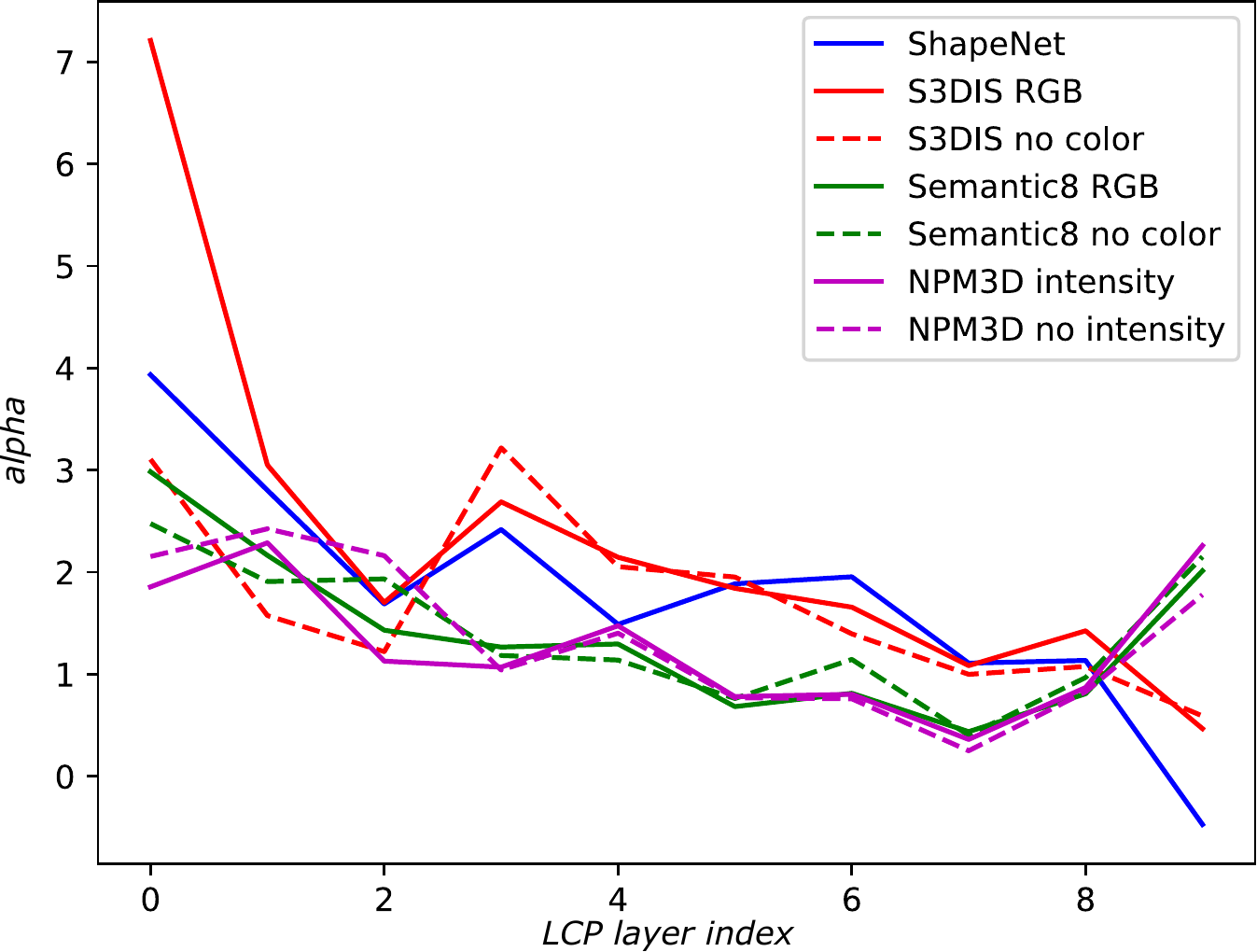}&
        \includegraphics[width=0.48\linewidth]{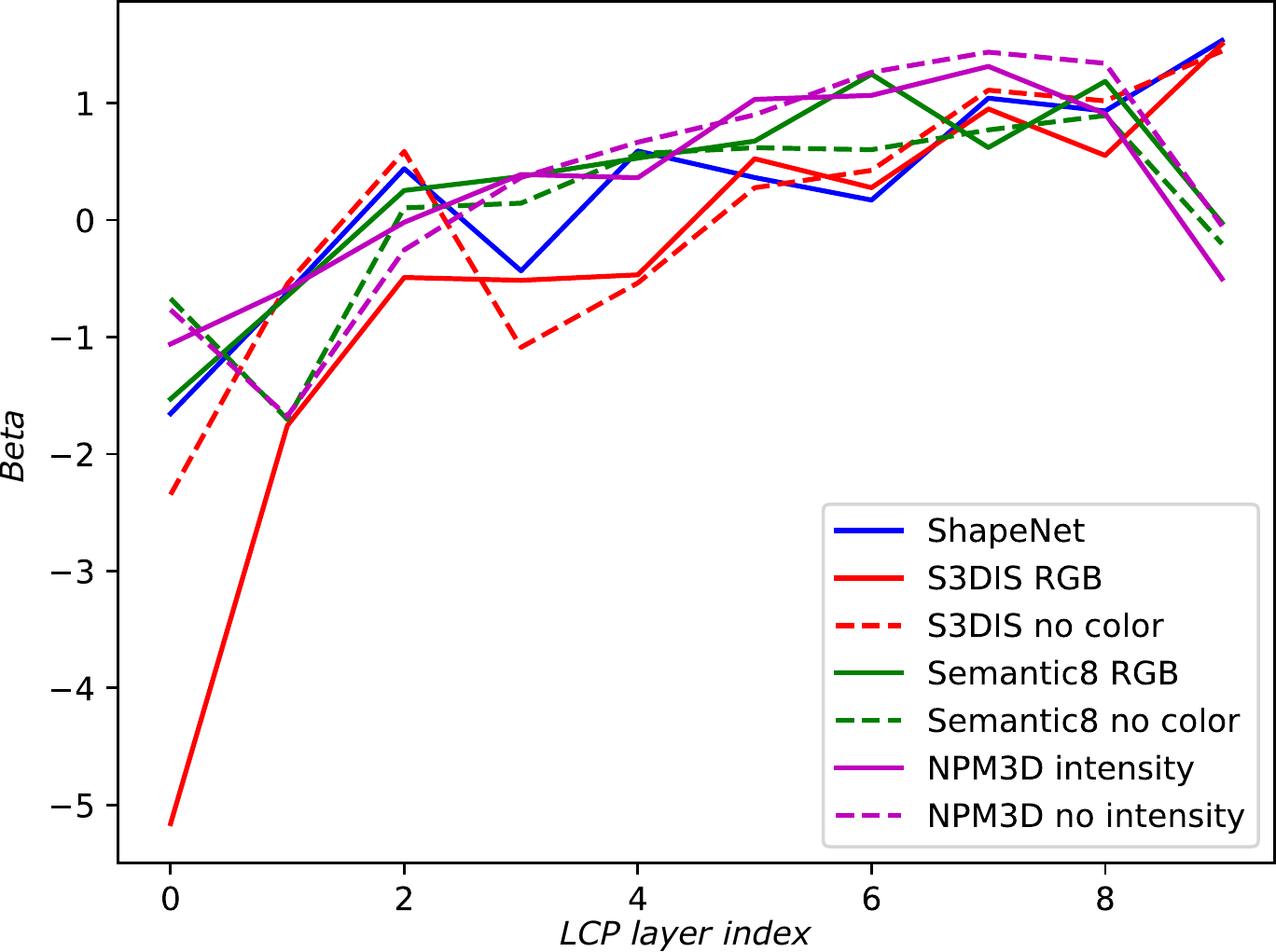} \\
        (b) $\alpha$ learned at each layer.& (c) $\beta$ learned at each layer.\\ 
        ~\\~\\
        \includegraphics[width=0.48\linewidth]{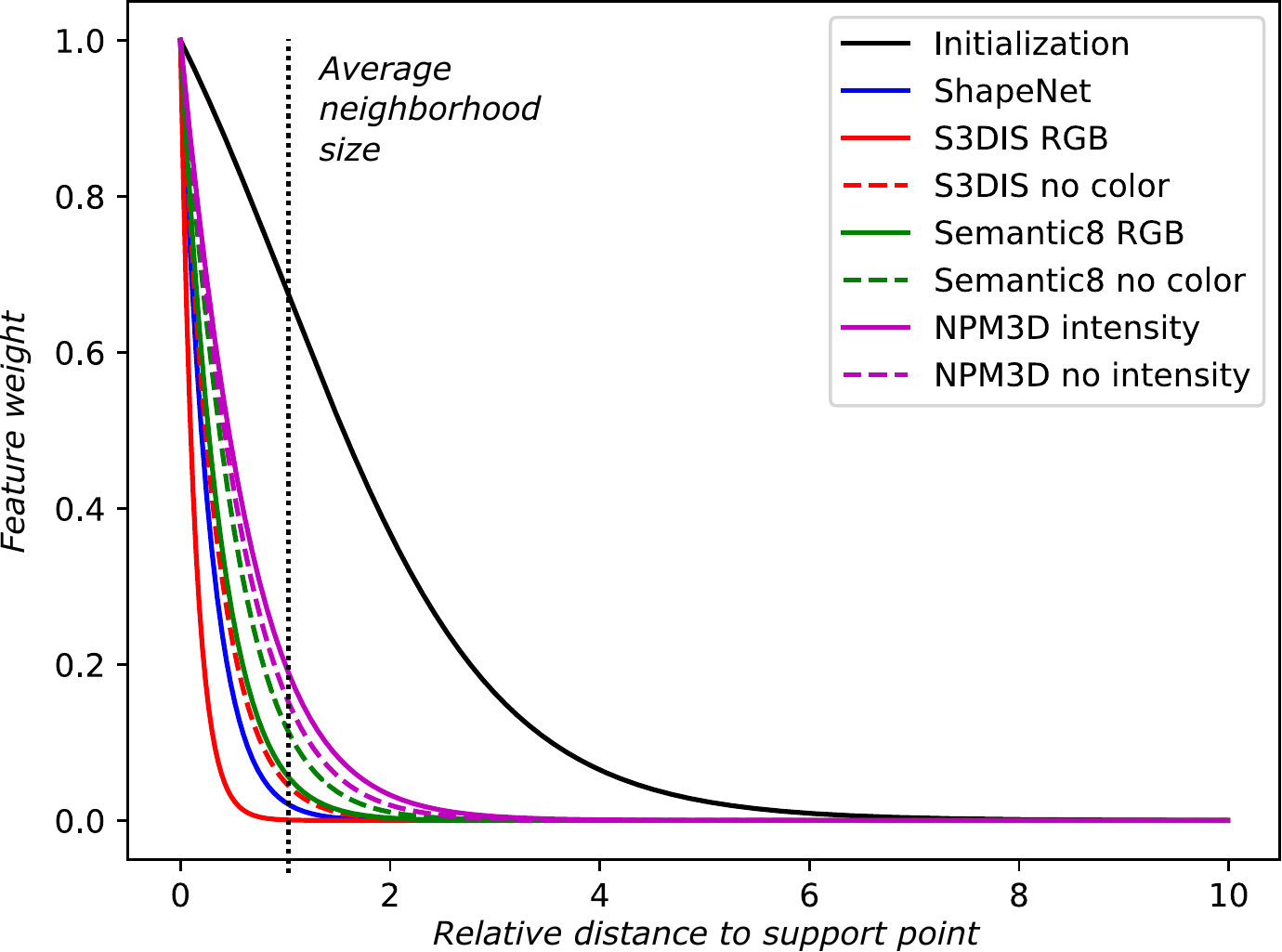}&
        \includegraphics[width=0.48\linewidth]{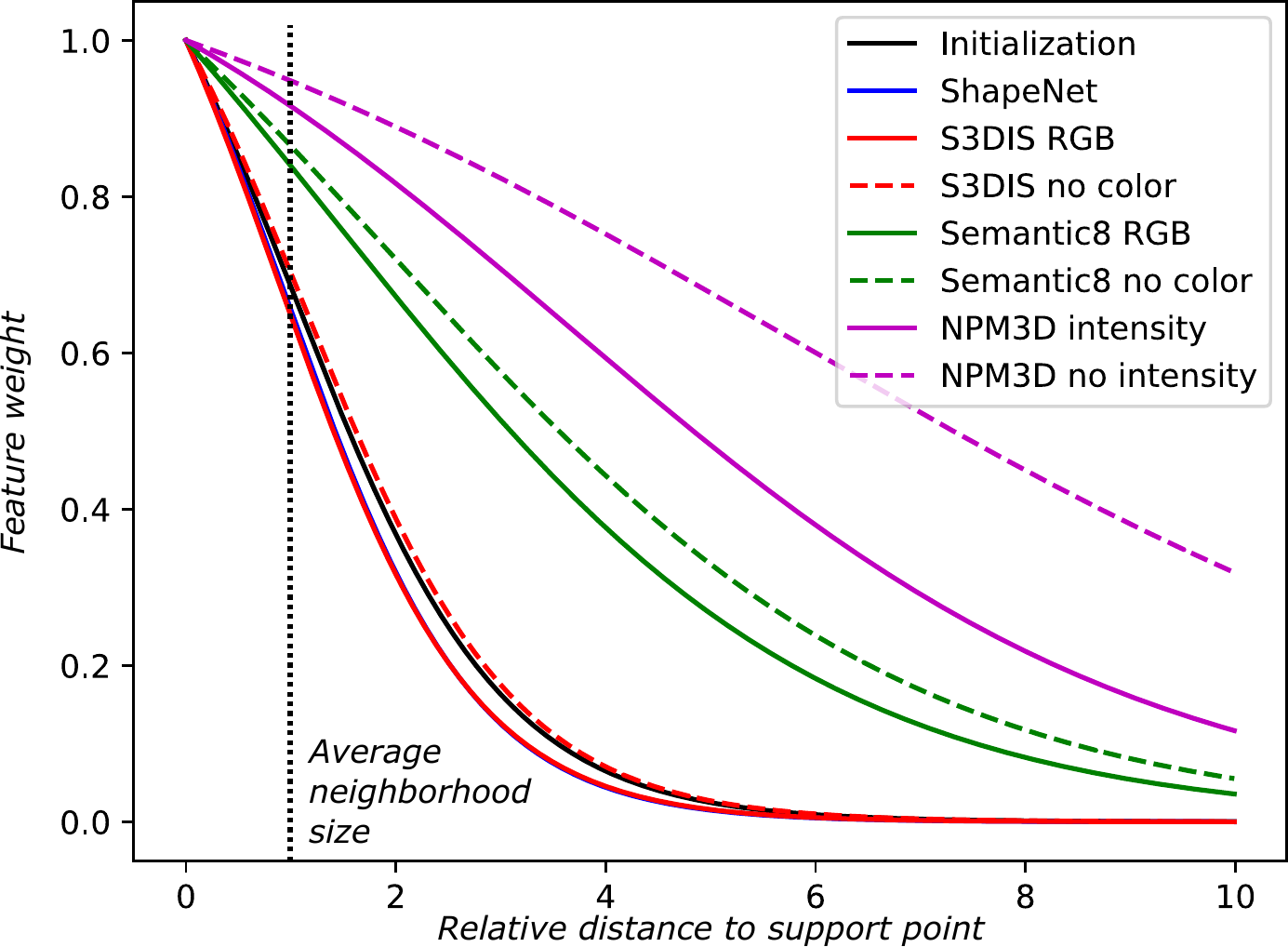} \\
        \multicolumn{2}{c}{Influence of the distance to support point on feature weight}\\
        (d) for the first FKAConv layer.  & (e) for the 7$^{\rm th}$ residual block. \\ 
    \end{tabular}
    \caption{Behavior of the learned neighborhood normalization and feature weighting across the segmentation network for various datasets.}
    \label{supp:fig:knn_rad}
\end{figure}

\subsubsection{Average neighborhood radius $r_{t}$.}

First, we study the estimated neighborhood size $r_{t}$ at each layer of the encoder after training.
This size is the averaged radius of the smallest sphere centered on the support point and encompassing the 16 nearest neighbors.
We present in Fig.\,\ref{supp:fig:knn_rad}\,(a) the evolution of this radius as a function of the layer's depth, for different datasets.

We observe that this radius is directly linked to the size of the bounding box of the input point cloud in the 3D space, i.e., to the size of the point cloud pillars (vertical infinite cylinders of diameter 8 meters for Semantic8 and NPM3D, and 2 meters for S3DIS) or to the size of the ShapeNet's CAD models.

\subsubsection{Weighting function.}

In Figs.\,\ref{supp:fig:knn_rad}\,(b-c), we plot the values of parameters $\alpha$ and $\beta$ as a function of the depth of the FKAConv layers.
First, we observe that these parameters, which are optimized with the network, take values that are different from the initial values that are set at the training initialization, i.e., $\alpha=1$ and $\beta=1$.
Second, the values of $\alpha$ and $\beta$ are similar after training on Semantic8 and NPM3D. This could be expected as both datasets share common characteristics: outdoor urban scenes, same pillar size.
Third, we observe the same global variations of the curves on all datasets: $\alpha$ tends to decrease with the depth while $\beta$ tends to increase. As a result, the transition of the gating function becomes wider in deeper layers, i.e., deeper layers tend to take into account more far away points.

We illustrate this phenomenon on Figs.\,\ref{supp:fig:knn_rad}\,(d-e), where we represent the weighting function after optimization for the first and for the seventh FKAConv layer of the network. For comparison purpose, we normalize the curves by setting the weight to 1 at distance 0. 
The black curve is the initial function, before optimization.

At the first layer, only the neighboring points that are very close to the support point are taken into consideration to estimate matrix $\mathbf{A}$.
We hypothesize that, in the absence of noise in data, a small neighborhood is sufficient to estimate local geometric features.

On the contrary, in the $7^{\rm th}$ layer, all the neighboring points are taken into consideration to compute $\mathbf{A}$.
At this stage, the number of support points is small and each point carries features that are discriminating for the task.
The network considers all available information, including points that are faraway from the support points.

\subsubsection{Influence on performance.}

\relax{
To quantify the impact of the neighborhood normalization and gating mechanism, we trained a segmentation network using five different configurations.
\begin{itemize}
    \item \textit{Baseline (no normalization)}. We do not normalize the neighborhood coordinates. The network sees different neighborhood sizes.
    \item \textit{Baseline (normalization to the unit ball)}. Each neighborhood is normalized into the unit sphere, regardless of its original size.
    \item \textit{Learned normalization + fixed gating at $r$}. The radius used for normalization is estimated using the proposed exponential moving average. The gating mechanism is replaced by hard-thresholding: features of all points outside the ball with the learned radius $r$ are set to zero.
    \item \textit{Learned normalization + fixed gating at $2r$}. Same as above but using a radius of $2r$ for hard-thresholding.
    \item \textit{Our approach}. The radius used for normalization and the gating function are learned as proposed.
\end{itemize}

\begin{table}[ht]
    \caption{Impact of the neighborhood normalization and gating on ShapeNet.}
    \label{supp:tab:knn_rad}
    \centering
    \vspace{-2mm}
    \begin{tabular}{l@{~}|@{~}c@{~}c}
        Method &  mcIoU & mIoU\\
        \hline
        Baseline (no normalization) & 84.3 & 85.3\\
        Baseline (normalization to unit ball) & 84.6     & 85.6 \\
        \hline
        Learned normalization + fixed gating & \\
        $\quad s_i = 1$ if $d(\mathbf{p}, \mathbf{q}) < r_t$, and $s_i = 0$ otherwise. & 83.8 & 84.8\\
        $\quad s_i = 1$ if $d(\mathbf{p}, \mathbf{q}) < 2r_t$, and $s_i = 0$ otherwise. & 84.0 & 85.2 \\
        \it\scriptsize $(\mathbf{p}$ denotes a point in the neighborhood of $\mathbf{q}$, and $r_t$ is\\
        \it\scriptsize the estimated radius in Equation (1).)\\
        \hline
        Our method (learned normalization and gating) & \textbf{84.8}     & \textbf{85.7} \\
        \hline
    \end{tabular}
\end{table}
The results, reported in Table~\ref{supp:tab:knn_rad}, show a slight improvement using our approach with respect to the baselines: the mean class intersection over union (mcIoU) increases by $0.2$ point and the instance average intersection over union (mIoU) by $0.1$ point.
This gain may seem small, but it is significant on this dataset as the performance are close to saturation in the leaderboard.

We also observe that the learned gating is an important factor of the success of our approach.
Fixed gating leads to a performance drop (see Table~\ref{supp:tab:knn_rad}).
This is due to the fact that hard-thresholding suppresses too much information, particularly in the late stages of the network where the neighborhoods are less regular and more subject to size variation.
}

\section{Qualitative results of segmentation}
\label{supp:sec:qualitative}

\begin{figure}[ht!]
    \centering
    \includegraphics[width=0.242\linewidth]{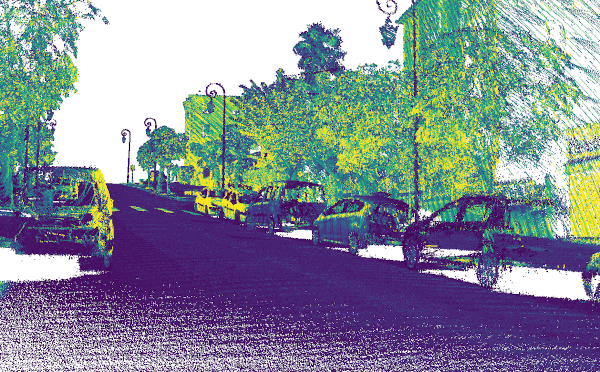}\,\includegraphics[width=0.242\linewidth]{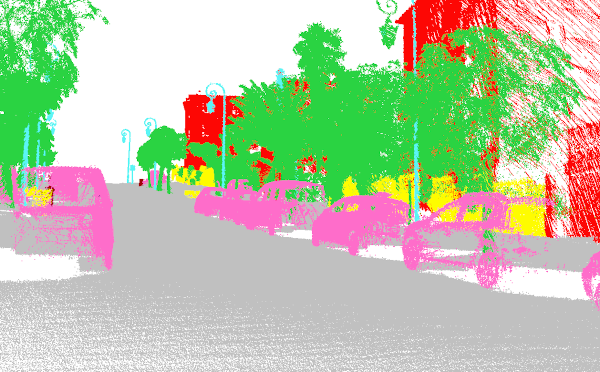}~~
    \includegraphics[width=0.242\linewidth]{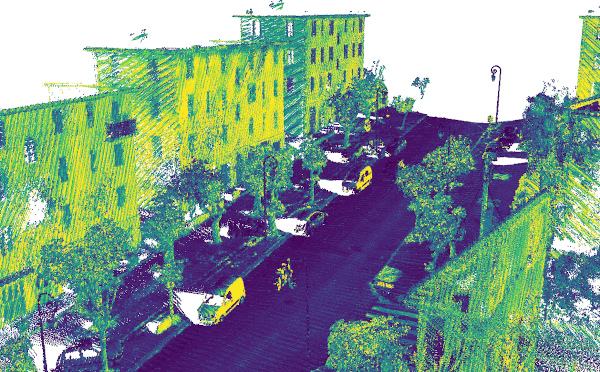}\,\includegraphics[width=0.242\linewidth]{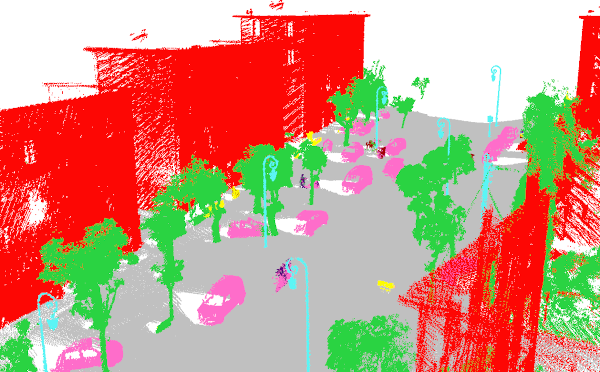}\\\vspace{2.0mm}
    \includegraphics[width=0.242\linewidth]{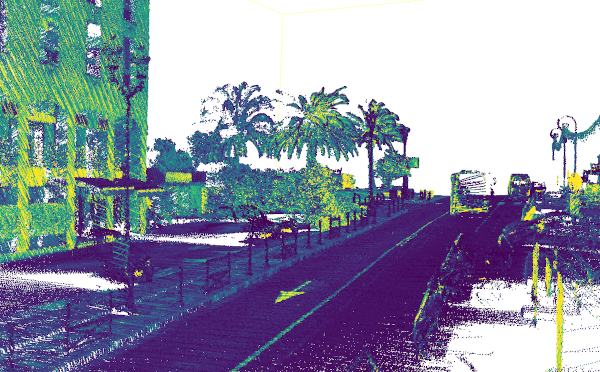}\,\includegraphics[width=0.242\linewidth]{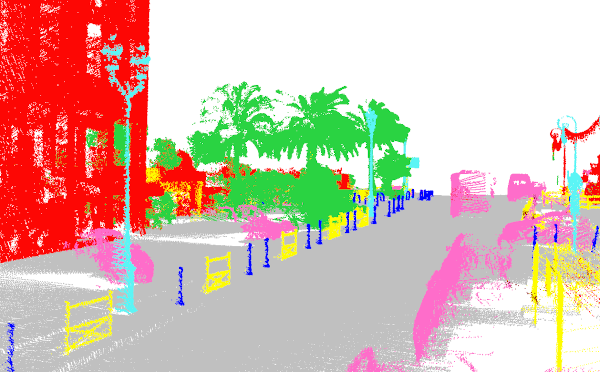}~~
    \includegraphics[width=0.242\linewidth]{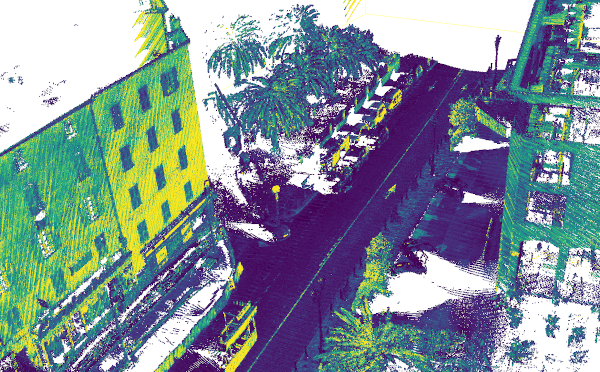}\,\includegraphics[width=0.242\linewidth]{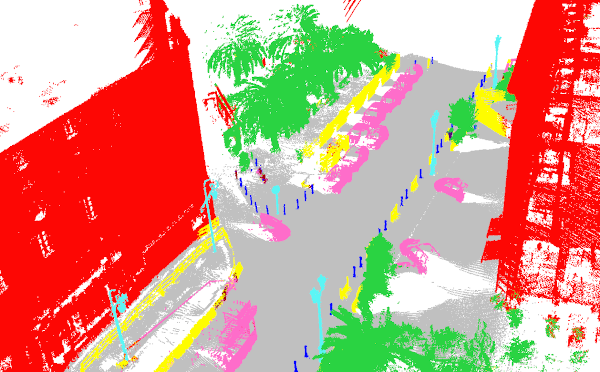}\\\vspace{2.0mm}
    \includegraphics[width=0.242\linewidth]{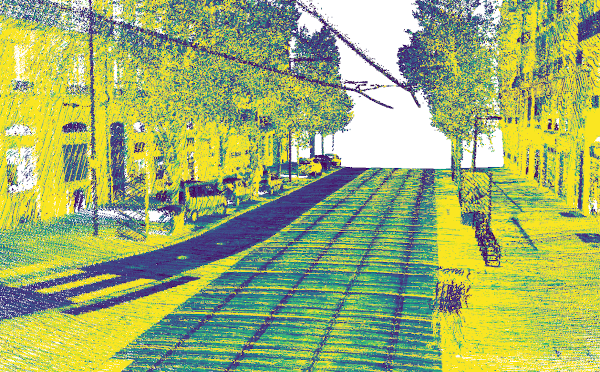}\,\includegraphics[width=0.242\linewidth]{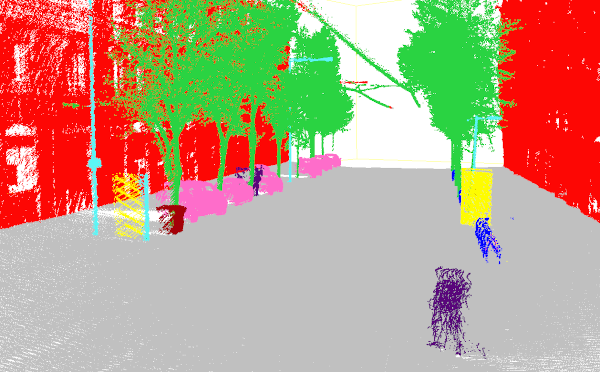}~~
    \includegraphics[width=0.242\linewidth]{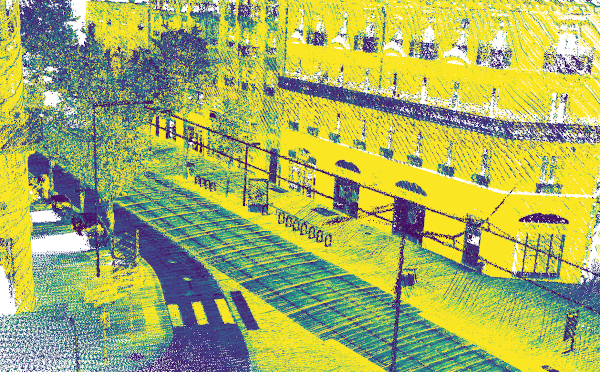}\,\includegraphics[width=0.242\linewidth]{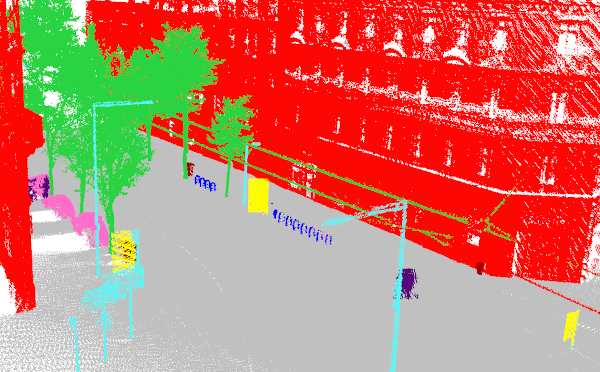}
    \caption{Visual results of our predictions on the test scenes of the NPM3D dataset. The input data is colored according to lidar intensity, from blue (low intensity) to red (high intensity) through green, yellow and orange. Colored, predicted, semantic segments are immediately on the right to lidar images. The segmentation results are obtained with the fusion model (intensity + geometry).}
    \label{fig:supp_npm3d_visu}
\end{figure}

\begin{figure}[p]
    \centering
    \includegraphics[width=0.24\linewidth]{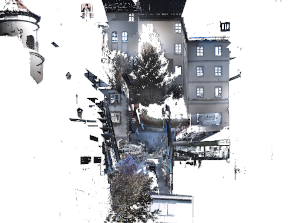}
    \includegraphics[width=0.24\linewidth]{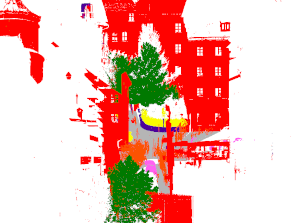}~~
    \includegraphics[width=0.24\linewidth]{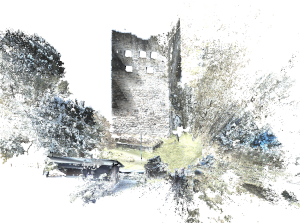}
    \includegraphics[width=0.24\linewidth]{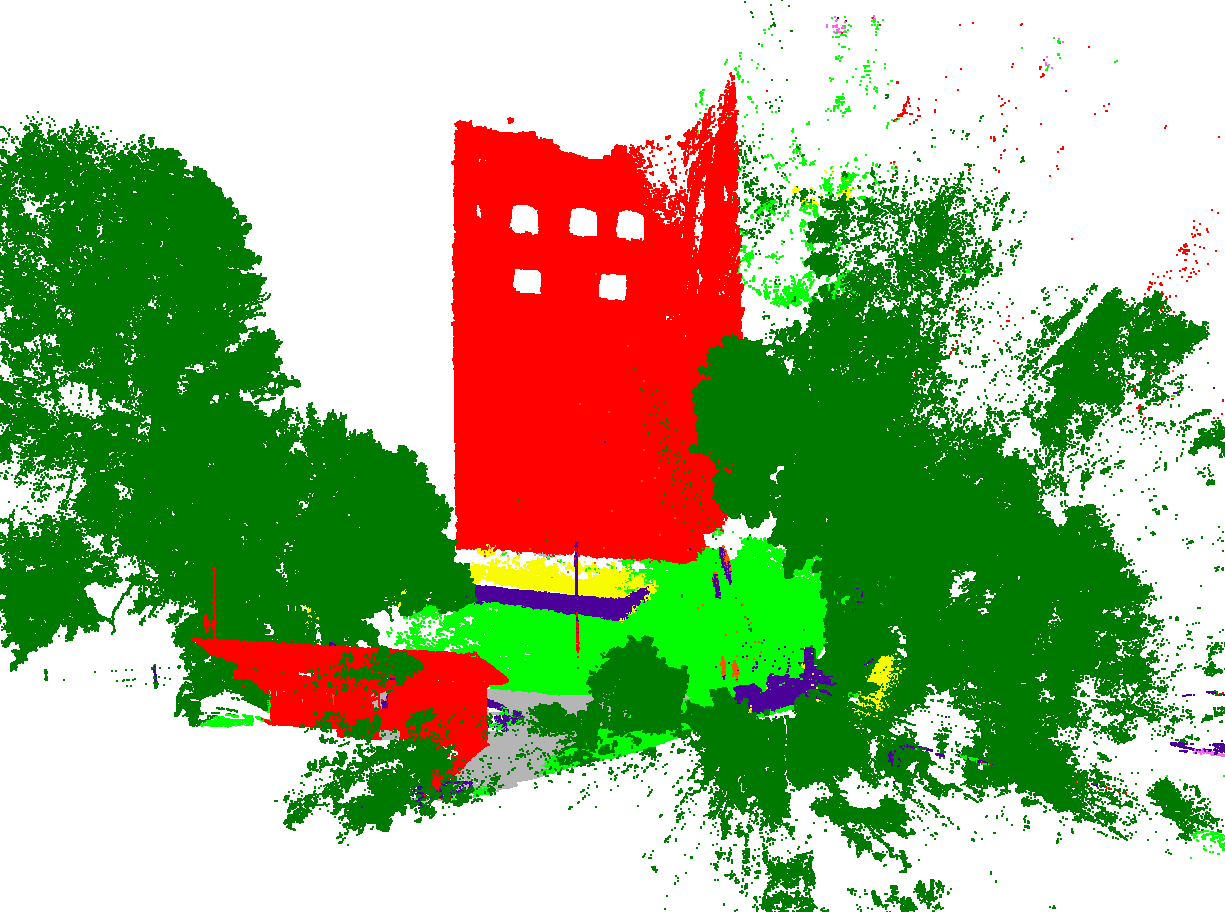}\\
    \includegraphics[width=0.24\linewidth]{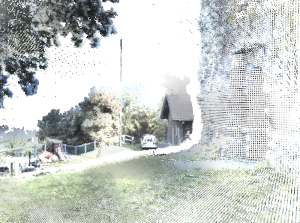}
    \includegraphics[width=0.24\linewidth]{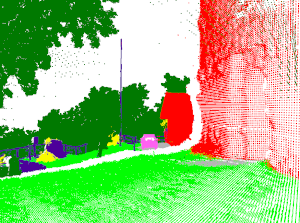}~~
    \includegraphics[width=0.24\linewidth]{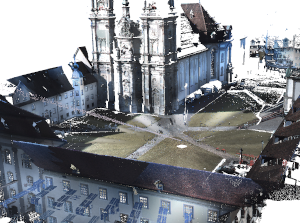}
    \includegraphics[width=0.24\linewidth]{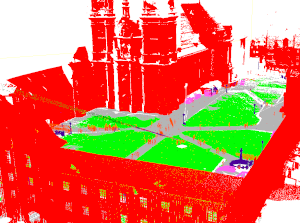}\\
    \includegraphics[width=0.24\linewidth]{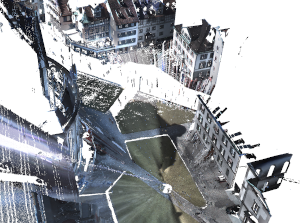}
    \includegraphics[width=0.24\linewidth]{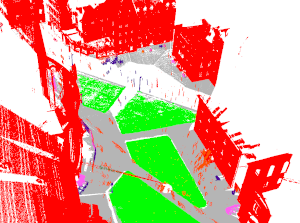}~~
    \includegraphics[width=0.24\linewidth]{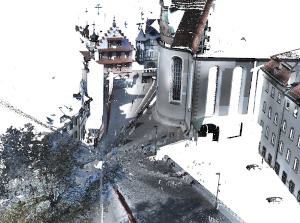}
    \includegraphics[width=0.24\linewidth]{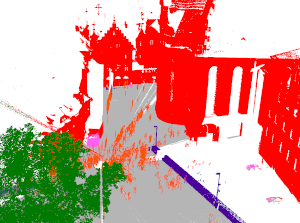}\\
    \includegraphics[width=0.24\linewidth]{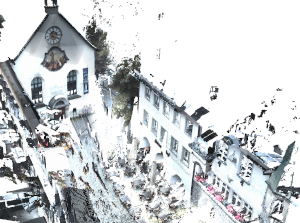}
    \includegraphics[width=0.24\linewidth]{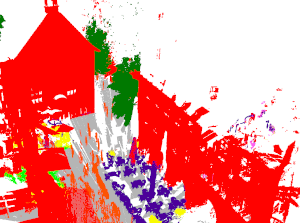}~~
    \includegraphics[width=0.24\linewidth]{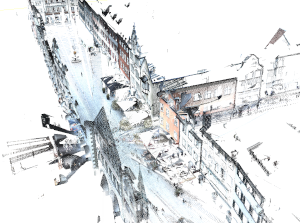}
    \includegraphics[width=0.24\linewidth]{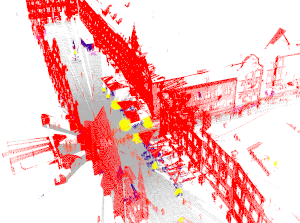}\\
    \includegraphics[width=0.24\linewidth]{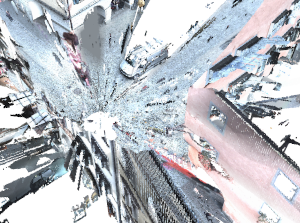}
    \includegraphics[width=0.24\linewidth]{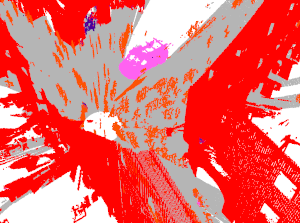}~~
    \includegraphics[width=0.24\linewidth]{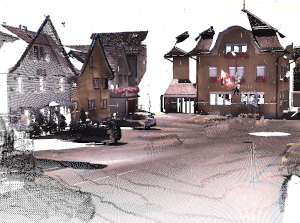}
    \includegraphics[width=0.24\linewidth]{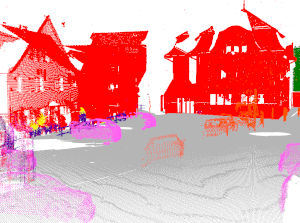}\\
    \includegraphics[width=0.24\linewidth]{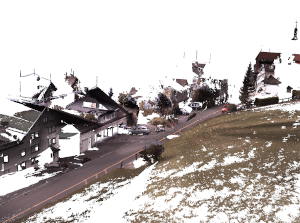}
    \includegraphics[width=0.24\linewidth]{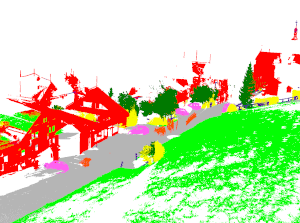}~~
    \includegraphics[width=0.24\linewidth]{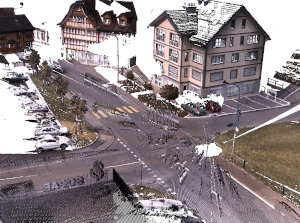}
    \includegraphics[width=0.24\linewidth]{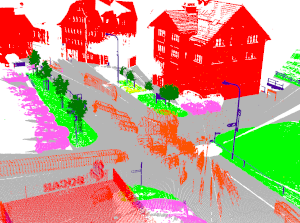}\\
    \includegraphics[width=0.24\linewidth]{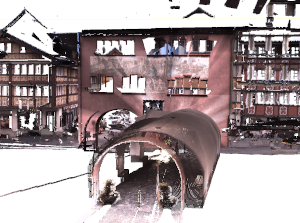}
    \includegraphics[width=0.24\linewidth]{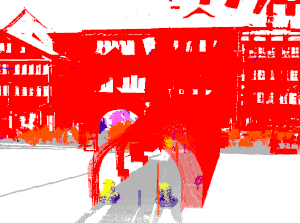}~~
    \includegraphics[width=0.24\linewidth]{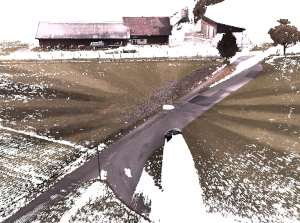}
    \includegraphics[width=0.24\linewidth]{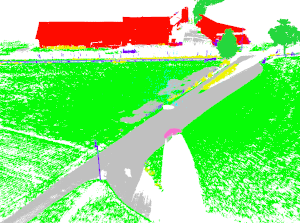}\\
    \includegraphics[width=0.24\linewidth]{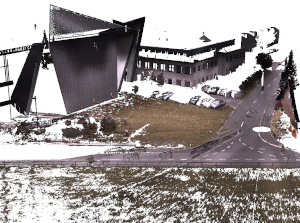}
    \includegraphics[width=0.24\linewidth]{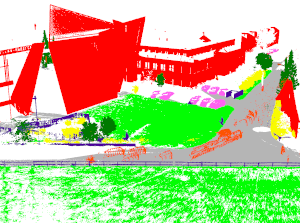}\\
    
    \caption{Visual results of our predictions on the 15 test scenes of the Semantic8 dataset. The segmentation results are obtained with the fusion model (RGB + geometry).}
    \label{fig:supp_sem8_visu}
\end{figure}

\begin{figure}[ht!]
    \centering
    \includegraphics[width=0.24\linewidth]{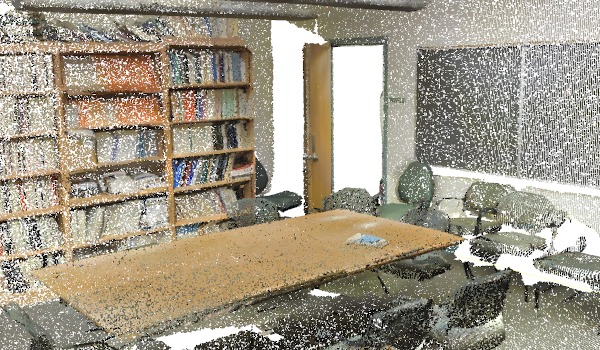}
    \includegraphics[width=0.24\linewidth]{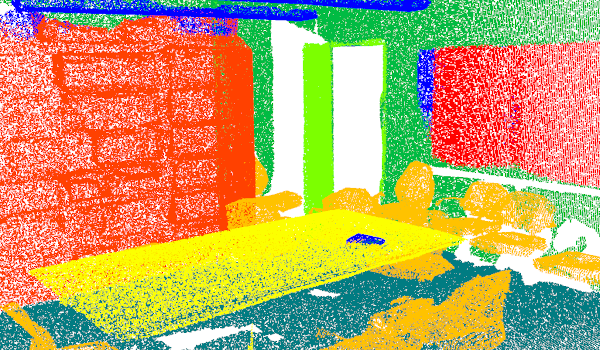}~~
    \includegraphics[width=0.24\linewidth]{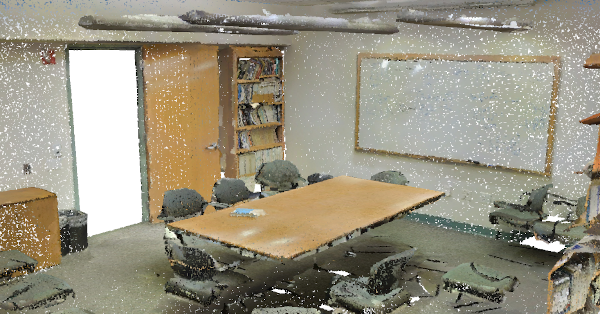}
    \includegraphics[width=0.24\linewidth]{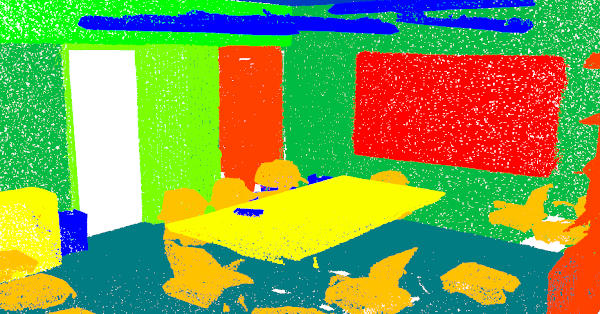}\\\vspace{2mm}
    \includegraphics[width=0.24\linewidth]{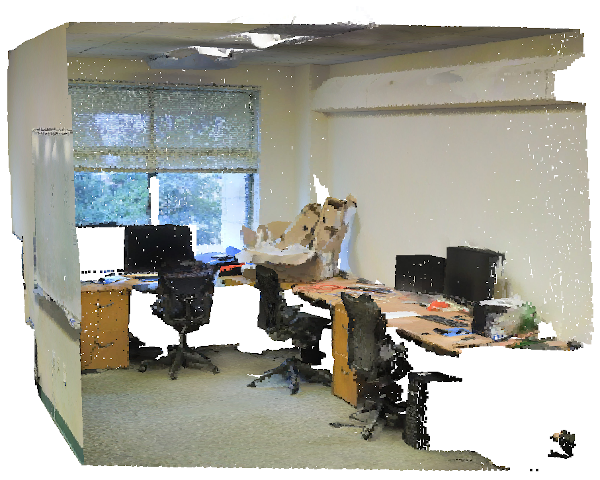}
    \includegraphics[width=0.24\linewidth]{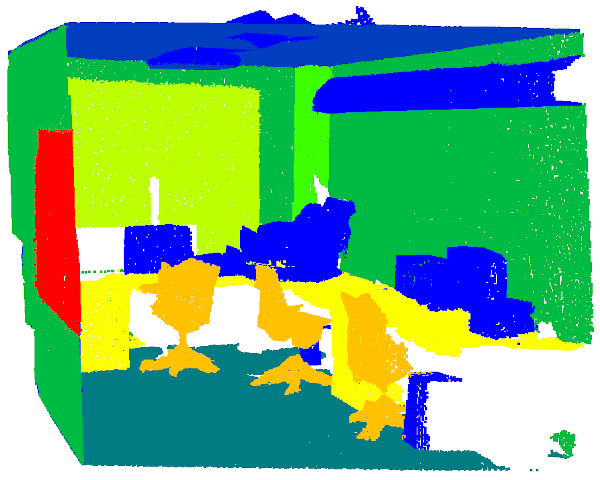}~~
    \includegraphics[width=0.24\linewidth]{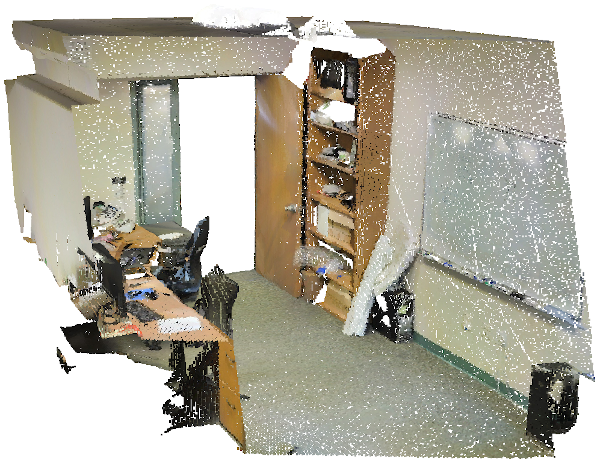}
    \includegraphics[width=0.24\linewidth]{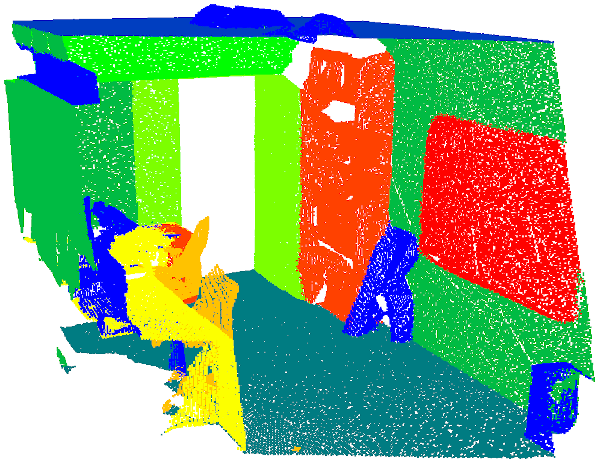}\\\vspace{2mm}
    \includegraphics[width=0.24\linewidth]{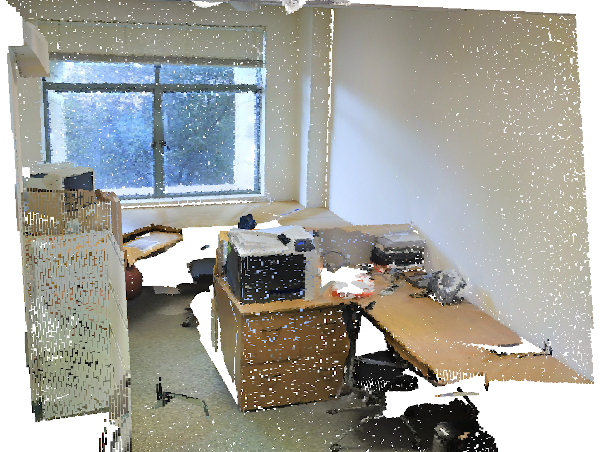}
    \includegraphics[width=0.24\linewidth]{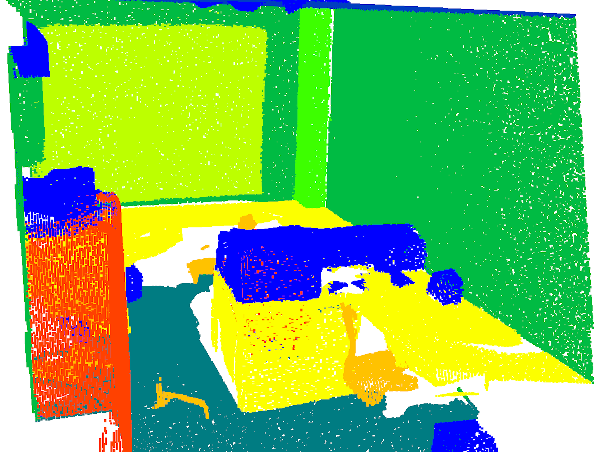}~~
    \includegraphics[width=0.24\linewidth]{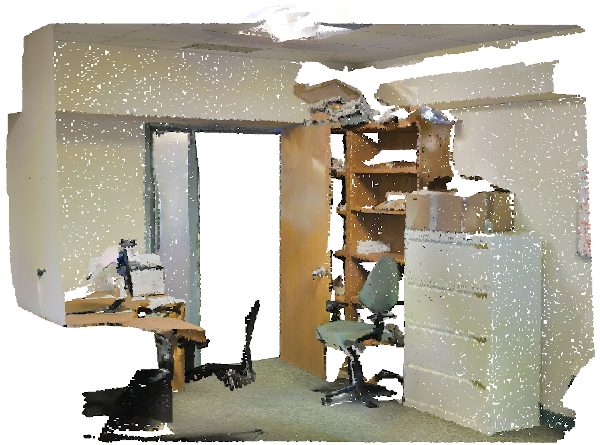}
    \includegraphics[width=0.24\linewidth]{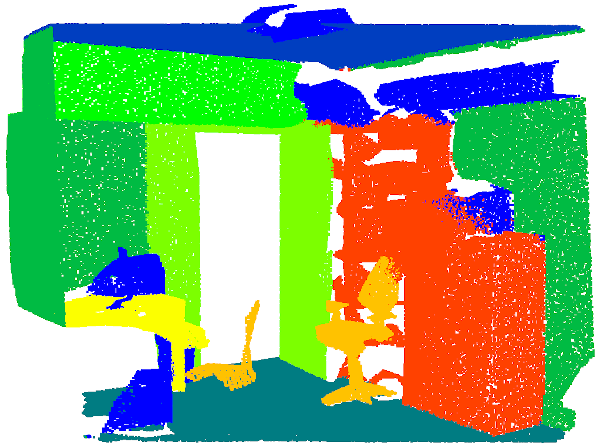}
    
    \caption{Visual results of our predictions on the S3DIS dataset obtained with the fusion model (RGB + geometry).}
    \label{fig:supp_s3dis_visu}
\end{figure}

Finally, we provide more visual results illustrating our semantic segmentation predictions on datasets NPM3D (Fig.\,~\ref{fig:supp_npm3d_visu}), Semantic8 (Fig.\,\ref{fig:supp_sem8_visu}) and S3DIS (Fig.\,\ref{fig:supp_s3dis_visu}).

